\documentclass[10pt,twocolumn,letterpaper]{article}

\usepackage{cvpr}
\usepackage{times}
\usepackage{epsfig}
\usepackage{graphicx}
\usepackage{amsmath}
\usepackage{amssymb}

\usepackage{epstopdf}
\usepackage[lined,boxed,commentsnumbered, ruled]{algorithm2e}

\usepackage[breaklinks=true,bookmarks=false]{hyperref}

\cvprfinalcopy 

\def\cvprPaperID{2374} 

\begin{document}

\title{Adaptive Object Detection Using Adjacency and Zoom Prediction}

\author{Yongxi Lu\\
University of California, San Diego\\
{\tt\small yol070@ucsd.edu}
\and
Tara Javidi\\
University of California, San Diego\\
{\tt\small tjavidi@ucsd.edu}
\and
Svetlana Lazebnik\\
University of Illinois at Urbana-Champaign\\
{\tt\small slazebni@illinois.edu}
}

\maketitle

\begin{abstract}
State-of-the-art object detection systems rely on an accurate set of region proposals. Several recent methods use a neural network architecture to hypothesize promising object locations. While these approaches are computationally efficient, they rely on fixed image regions as anchors for predictions. In this paper we propose to use a search strategy that adaptively directs computational resources to sub-regions likely to contain objects. Compared to methods based on fixed anchor locations, our approach naturally adapts to cases where object instances are sparse and small. Our approach is comparable in terms of accuracy to the state-of-the-art Faster R-CNN approach while using two orders of magnitude fewer anchors on average. Code is publicly available. 

\end{abstract}

\section{Introduction}
Object detection is an important computer vision problem for its intriguing challenges and large variety of applications. Significant recent progress in this area has been achieved by incorporating deep convolutional neural networks (DCNN) \cite{krizhevsky2012imagenet} into object detection systems \cite{erhan2014scalable,gidaris2015object,girshick2014rich,he2014spatial,liang2014computational,simonyan2014very,szegedy2013deep}. 

An object detection algorithm with state-of-the-art accuracy typically has the following two-step cascade: a set of class-independent region proposals are hypothesized and are then used as input to a detector that gives each region a class label. The role of region proposals is to reduce the complexity through limiting the number of regions that need be evaluated by the detector. However, with recently introduced techniques that enable sharing of convolutional features \cite{girshick2015fast,he2014spatial}, traditional region proposal algorithms such as selective search \cite{uijlings2013selective} and EdgeBoxes \cite{zitnick2014edge} become the bottleneck of the detection pipeline. 

An emerging class of efficient region proposal methods are based on end-to-end trained deep neural networks \cite{erhan2014scalable,ren2015faster}. The common idea in these approaches is to train a class-independent regressor on a small set of pre-defined anchor regions. More specifically, each anchor region is assigned the task of deciding whether an object is in its neighborhood (in terms of center location, scale and aspect ratio), and predicting a bounding box for that object through regression if that is the case. The design of anchors differs for each method. For example, MultiBox \cite{erhan2014scalable} uses 800 anchors from clustering, YOLO \cite{redmon2015you} uses a non-overlapping 7 by 7 grid, RPN \cite{ren2015faster} uses overlapping sliding windows. In these prior works the test-time anchors are not adaptive to the actual content of the images, thus to further improve accuracy for detecting small object instances a denser grid of anchors is required for all images, resulting in longer test time and a more complex network model. 

\begin{figure*}[t]
\begin{center}
   \includegraphics[width=3in, height=2in]{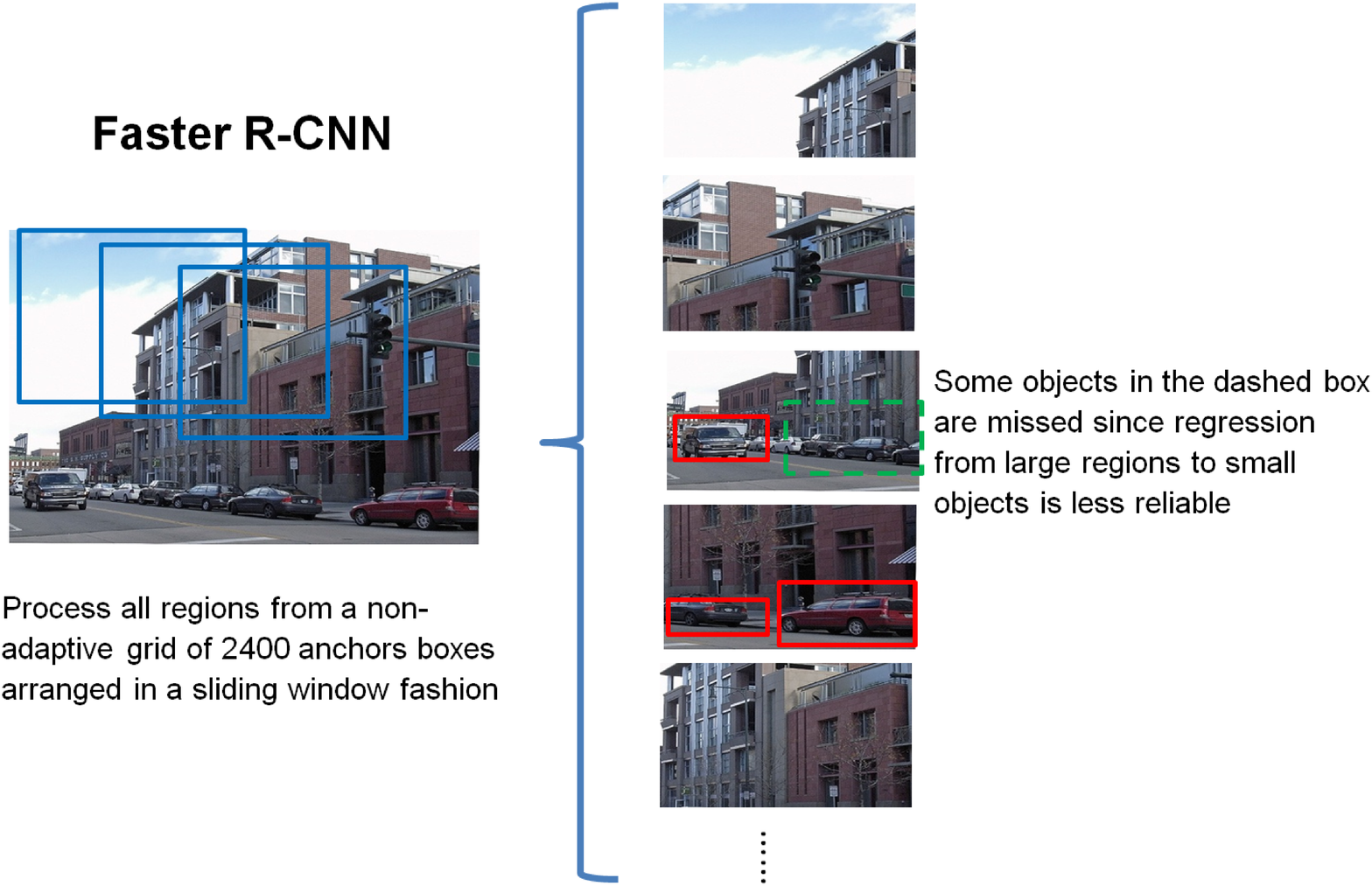}
   \hspace{0.5cm}
   \includegraphics[height=2in]{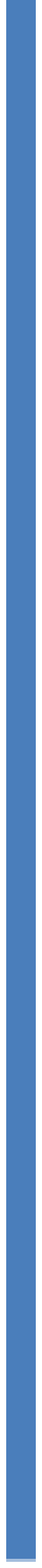} 
   \hspace{0.5cm}  
   \includegraphics[width=3in, height=2in]{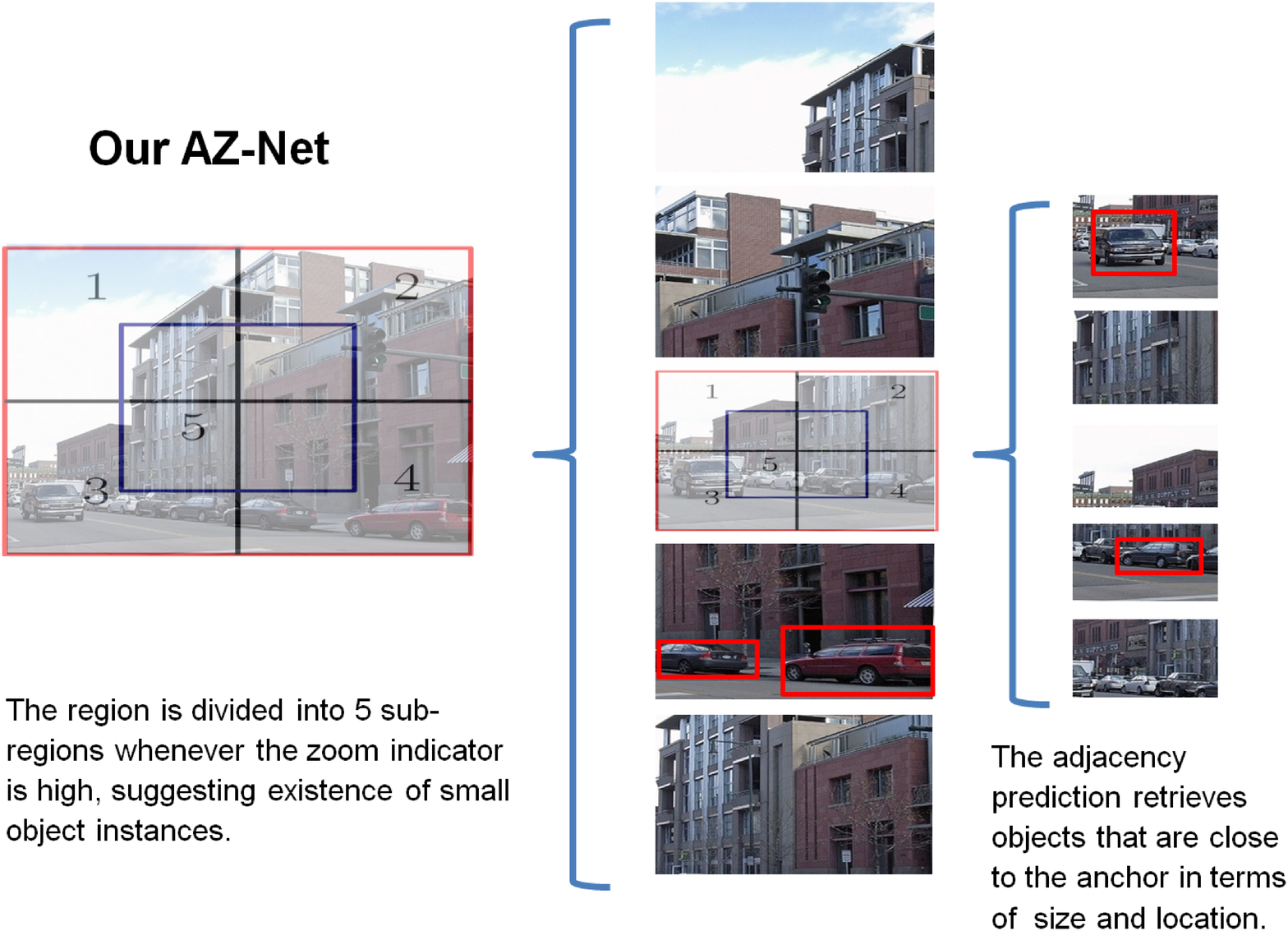}
\end{center}
   \caption{Comparison of our proposed adaptive search algorithm with the non-adaptive RPN method. The red boxes show region proposals from adjacency predictions. Note that for small objects, RPN is forced to perform regression from much larger anchors, while our AZ-Net approach can adaptively use features from small regions. }
\label{fig:pipeline}
\end{figure*}

\begin{figure}[t]
\begin{center}
   \includegraphics[width=1in]{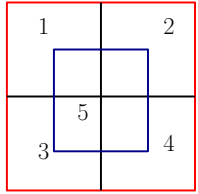}
\end{center}
   \caption{As illustrated, a given region is divided into 5 sub-regions (numbered). Each of these sub-regions is recursively divided if its zoom indicator is above a threshold. }
\label{fig:divide}
\end{figure}

We alternatively consider the following adaptive search strategy. Instead of fixing {\em a priori} a set of anchor regions, our algorithm starts with the entire image. It then recursively divides the image into sub-regions (see Figure \ref{fig:divide}) until it decides that a given region is unlikely to enclose any small objects. The regions that are visited in the process effectively serve as anchors that are assigned the task of predicting bounding boxes for objects nearby. A salient feature of our algorithm is that the decision of whether to divide a region further is based on features extracted from that particular region. As a result, the generation of the set of anchor regions is conditioned on the image content. For an image with only a few small objects most regions are pruned early in the search, leaving a few small anchor regions near the objects. For images that contain exclusively large instances, our approach gracefully falls back to existing methods that rely on a small number of large anchor regions. In this manner, our algorithm adaptively directs its computational resources to regions that are likely to contain objects. Figure \ref{fig:pipeline} compares our algorithm with RPN. 

To support our adaptive search algorithm, we train a deep neural network we call {\em Adjacency and Zoom Network (AZ-Net)}. Given an input anchor region, the AZ-Net outputs a scalar \textit{zoom indicator} which is used to decide whether to further zoom into (divide) the region and a set of bounding boxes with confidence scores, or \textit{adjacency predictions}. The adjacency predictions with high confidence scores are then used as region proposals for a subsequent object detector. The network is applied recursively starting from the whole image to generate an adaptive set of proposals. 

To intuitively motivate the design of our network, consider a situation in which one needs to perform a quick search for a car. A good strategy is to first look for larger structures that could provide evidence for existence of smaller structures in related categories. A search agent could, for example, look for roads and use that to reason about where cars should be. Once the search nears the car, one could use the fact that seeing certain parts is highly predictive of the spatial support of the whole. For instance, the wheels provide strong evidence for a tight box of the car. In our design, the zoom indicator mimics the process of searching for larger structures, while the adjacency predictions mimic the process of neighborhood inference. 

To validate this design we extensively evaluate our algorithm on Pascal VOC 2007 \cite{pascal-voc-2007} with fine-grained analysis. We also report baseline results on the recently introduced MSCOCO \cite{lin2014microsoft} dataset. Our algorithm achieves detection mAP that is close to state-of-the-art methods at a fast frame rate. Code has been made publicly available at \url{https://github.com/luyongxi/az-net}.

In summary, we make the following contributions:
\begin{itemize}
\item We design a search strategy for object detection that adaptively focuses computational resources on image regions that contain objects. 
\item We evaluate our approach on Pascal VOC 2007 and MSCOCO datasets and demonstrate it is comparable to Fast R-CNN and Faster R-CNN with fewer anchor and proposal regions. 
\item We provide a fine-grained analysis that shows intriguing features of our approach. Namely, our proposal strategy has better recall for higher intersection-over-union thresholds, higher recall for smaller numbers of top proposals, and for smaller object instances.
\end{itemize}

This paper is organized as follows. In section \ref{sec:related} we survey existing literature highlighting the novelty of our approach. In Section \ref{sec:design} we introduce the design of our algorithm. Section \ref{sec:exp} presents an empirical comparison to existing object detection methods on standard evaluation benchmarks, and Section \ref{sec:con} discusses possible future directions. 

\section{Previous Work}
\label{sec:related}
Lampert \etal \cite{lampert2009efficient} first proposed an adaptive branch-and-bound approach. More recently, Gonzeles-Garcia \etal \cite{gonzalez2015active}, Caicedo and Lazebnik \cite{caicedoactive}, and Yoo \etal \cite{yoo2015attentionnet} explored active object detection with DCNN features. While these approaches show the promise of using an adaptive algorithm for object detection, their detectors are class-wise and their methods cannot achieve competitive accuracy. Our approach, on the other hand, is multi-class and is comparable to state-of-the-art approaches in both accuracy and test speed. 

The idea of using spatial context has been previously explored in the literature. Previous work by Torralba \etal \cite{torralba2006contextual} used a biologically inspired visual attention model \cite{borji2013state}, but our focus is on efficient engineering design. Divvala \etal \cite{divvala2009empirical} evaluated the use of context for localization, but their empirical study was performed on hand-crafted features and needs to be reexamined in combination with more accurate recent approaches. 

Our method is closely related to recent approaches that use anchor regions for proposal generation or detection. For example, Erhan \etal \cite{erhan2014scalable} use 800 data-driven anchors for region proposals and Redmon \etal \cite{redmon2015you} use a fixed grid of 49 non-overlapping regions to provide class-wise detections. The former has the concern that these anchors could overfit the data, while the latter cannot achieve state-of-the-art performance without model ensembles. Our work is most related to the recent work by Ren \etal \cite{ren2015faster}, which uses a set of heuristically designed 2400 overlapping anchor regions. Our approach uses a similar regression technique to predict multiple bounding boxes from an anchor region. However, our anchor regions are generated adaptively, making them intrinsically more efficient. In particular, we show that it is possible to detect small object instances in the scene without an excessive number of anchor regions. We propose to grow a tree of finer-grained anchor regions based on local image evidence, and design the regression model strategically on top of it. We extensively compare the output of our method against \cite{ren2015faster} in our experimental section and show the unique advantages of our approach. 

This paper is a follow-up to the work published in the 53rd Annual Allerton Conference \cite{lu2015efficient}. Here, we introduce a substantially improved algorithm and add extensive evaluations on standard benchmarks.

\begin{figure}[t]
\begin{center}
      \includegraphics[height=0.7in, width=0.8in]{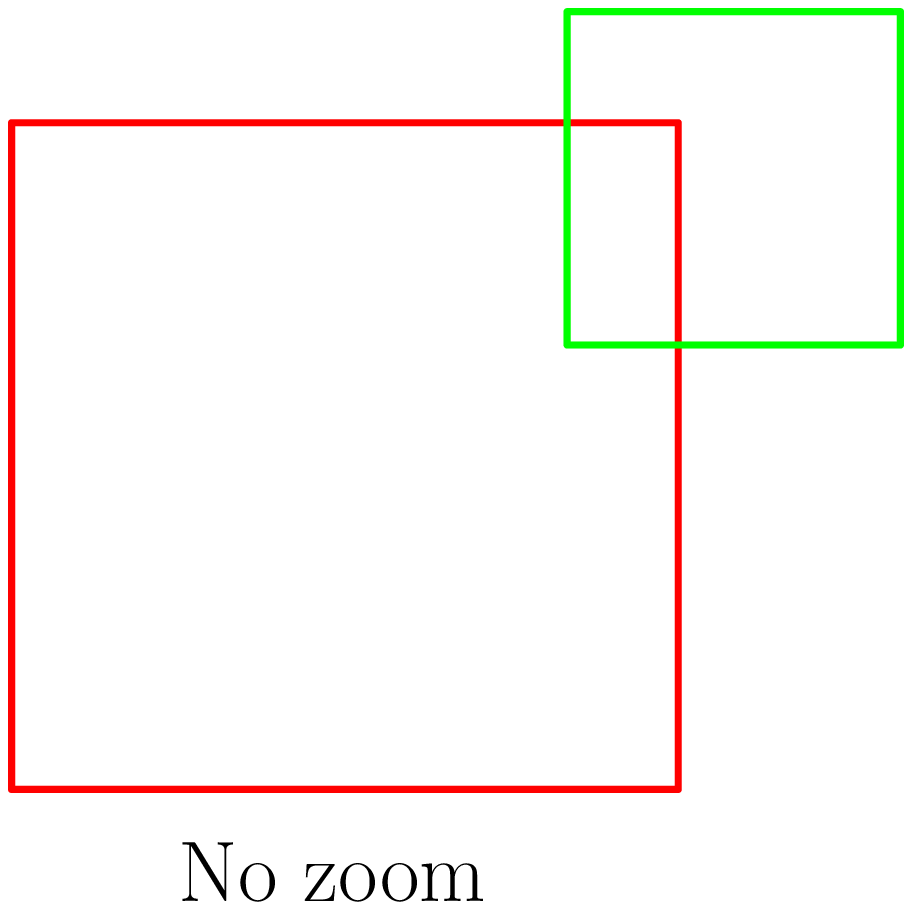}
      \hspace{0.5cm}
      \includegraphics[height=0.7in, width=0.8in]{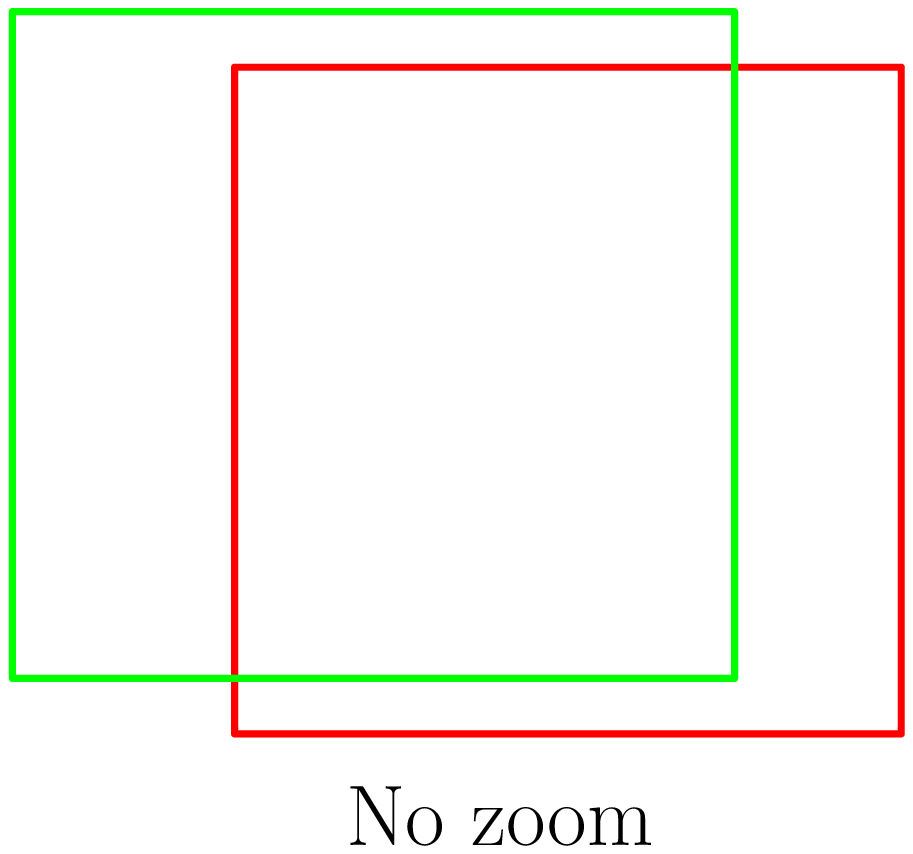}
      \hspace{0.5cm}
      \includegraphics[height=0.7in, width=1in]{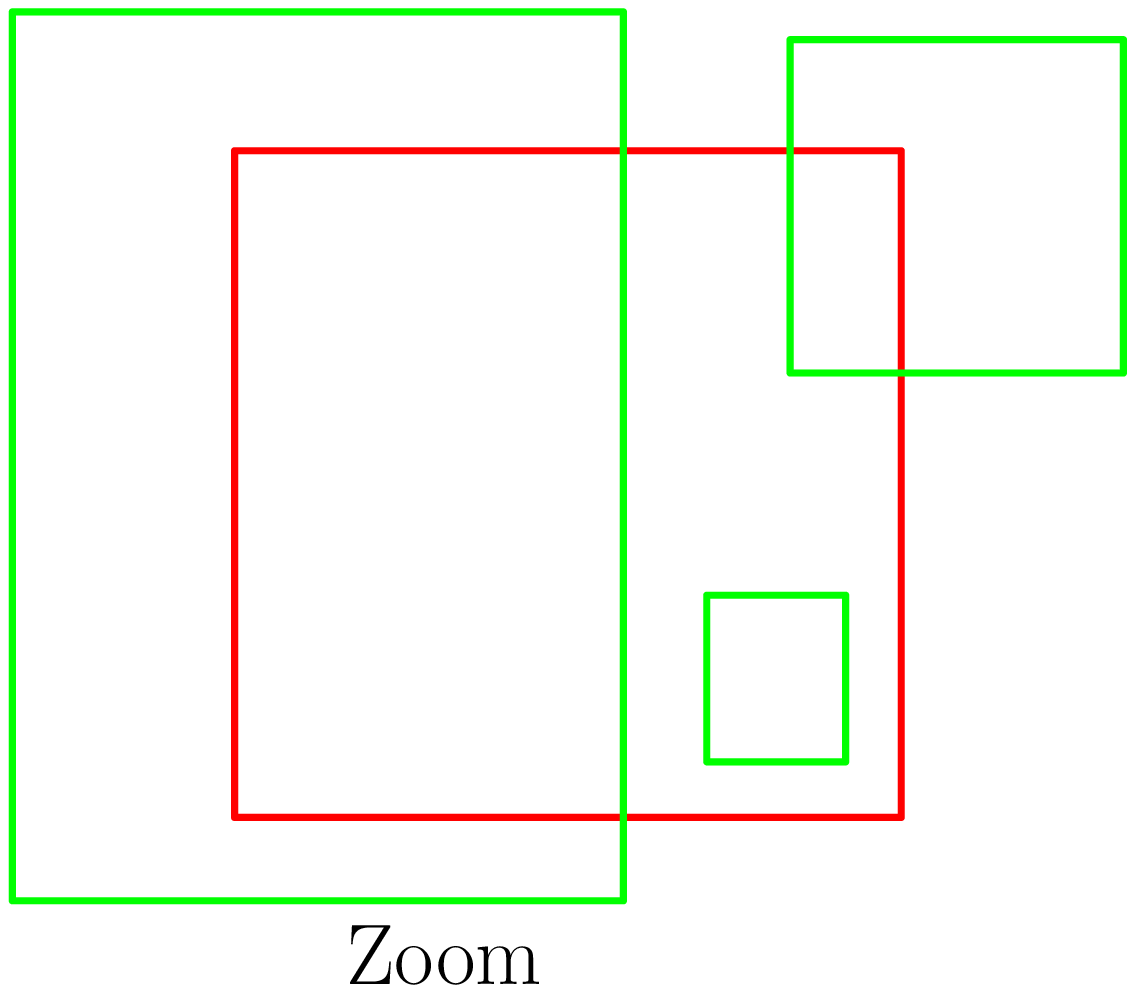}        
\end{center}
   \caption{Illustration of desired zoom indicator for common situations. The green boxes are objects, and the red boxes are regions. Left: the object is small but it is mostly outside the region -- there is no gain in zooming in. Middle: the object is mostly inside but its size is large relative to the region -- there is no gain in zooming in. Right: there is a small object that is completely inside the region. In this case further division of the region greatly increases the chance of detection for that object. }
\label{fig:zoom}
\end{figure}

\begin{figure}[t]
\begin{center}
      \includegraphics[height=0.7in, width=0.7in]{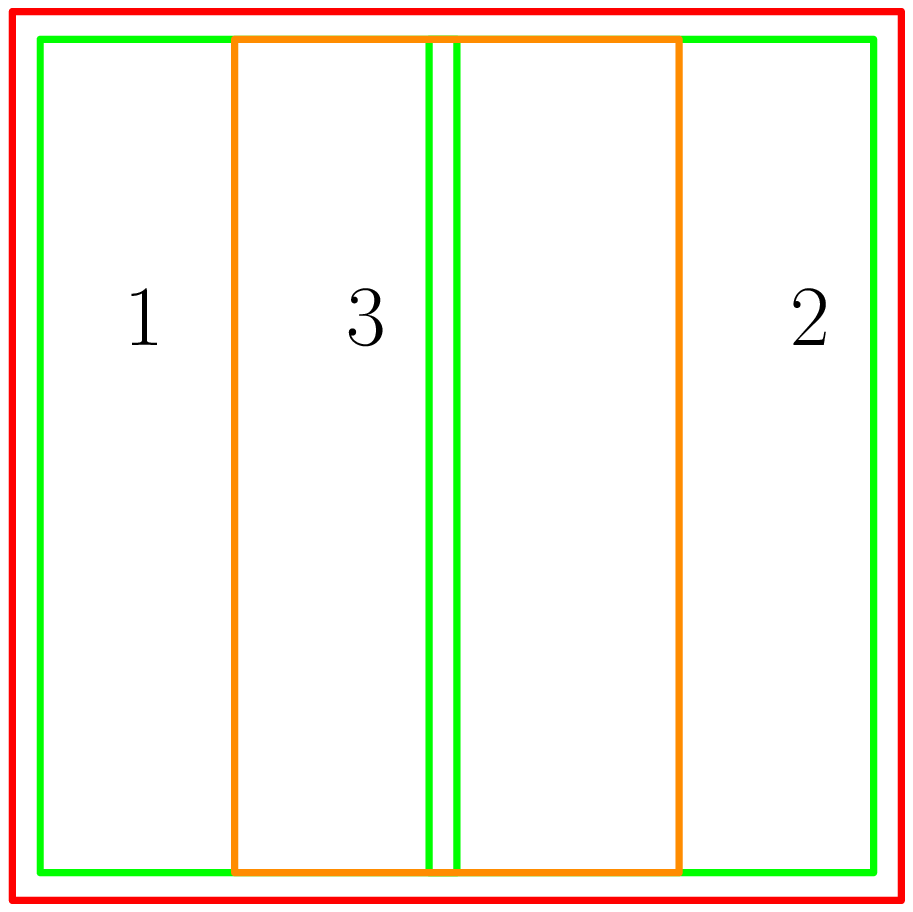}
      \hspace{0.5cm}
      \includegraphics[height=0.7in, width=0.7in]{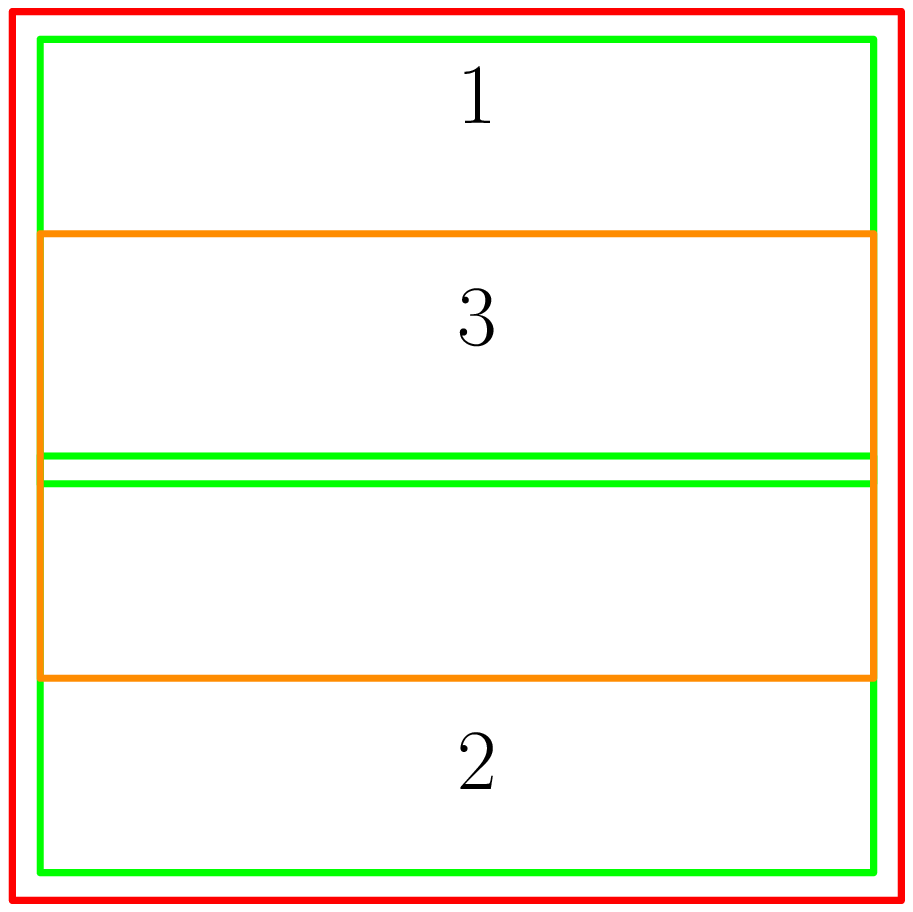}
      \hspace{0.5cm}
      \includegraphics[height=0.7in, width=0.7in]{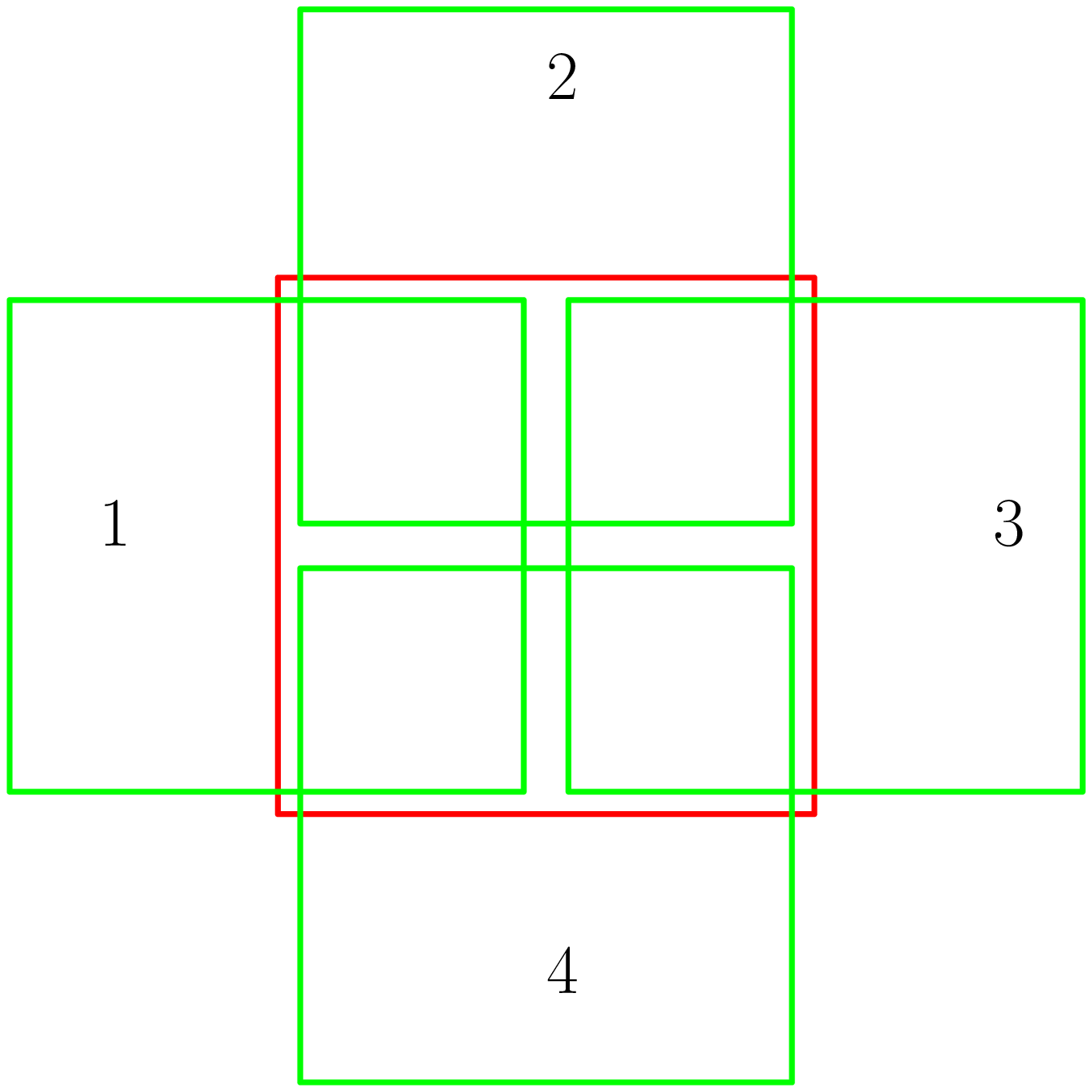}      
\end{center}
      \caption{Illustration of sub-region priors. From left to right: vertical stripes, horizontal stripes, neighboring squares. The red rectangular box is the image. In the figure the numbered regions are template sub-regions. The gaps between sub-regions are exaggerated for better visualization. The vertical stripes are used to detect tall objects, the horizontal stripes are used to detect fat objects, while the neighboring squares are used to detect objects that fall in the gaps between anchor regions generated in the search process. }
\label{fig:sub-region}
\end{figure}

\begin{algorithm*}\label{alg:region-propose}
 \SetAlgoLined

\KwData{Input image $x$ (the whole image region $b_x$). $Y_k$ is the region proposed at step $k$. $Y^k$ are the accumulated region proposals up to step $k$. $Z_k$ are the regions to further zoom in to at step $k$. $B_k$ are anchor regions at step $k$.}
\KwResult{Region proposals at termination $Y^K$.}
Initialization: $B_0 \leftarrow \{b_x\}$. $Y^0 \leftarrow \emptyset$, $k \leftarrow 0$ \\
\While{($B_k$ is not an empty set)}{
Initialize $Y_k$ and $Z_k$ as empty sets. \\
\ForEach{$b \in B_k$}{Compute adjacency predictions $A_b$ and the zoom indicator $z_b$ using AZ-Net.\\
Include all $a \in A_b$ with high confidence scores into $Y_k$. \\
Include $b$ into $Z_k$ if $z_b$ is above threshold. \\
}
$Y^k \leftarrow Y^{k-1} \cup Y_k$ \\
$B_{k+1} \leftarrow \mbox{Divide-Regions}(Z_{k})$ \\
$k \leftarrow k+1$}
$K \leftarrow k-1$
\caption{Adaptive search with AZ-Net.}
\end{algorithm*}

\section{Design of the Algorithm}
\label{sec:design}

\subsection{Overview of the Adaptive Search}
Our object detection algorithm consists of two steps. In step 1, a set of class-independent region proposals are generated using Adaptive Search with AZ-Net (see Algorithm \ref{alg:region-propose}). In step 2, an object detector evaluates each region proposed in step 1 to provide class-wise detections. In our experiments the detector is Fast R-CNN. 

Our focus is on improving step 1. We consider a recursive search strategy, starting from the entire image as the root region. For any region encountered in the search procedure, the algorithm extracts features from this region to compute the zoom indicator and the adjacency predictions. The adjacency predictions with confidence scores above a threshold are included in the set of output region proposals. If the zoom indicator is above a threshold, this indicates that the current region is likely to contain small objects. To detect these embedded small objects, the current region is divided into sub-regions in the manner shown in Figure \ref{fig:divide}.  Each of these sub-regions is then recursively processed in the same manner as its parent region, until either its area or its zoom indicator is too small. Figure \ref{fig:pipeline} illustrates this procedure.

In the following section, we discuss the design of the zoom indicator and adjacency prediction. 

\begin{figure}[t]
\begin{center}
\includegraphics[width=3.2in]{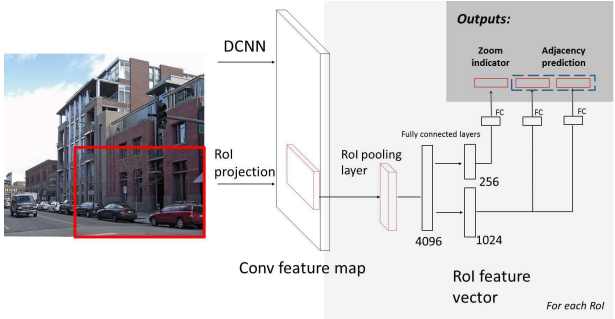}  
\end{center}
   \caption{Illustration of the AZ-Net architecture.}
\label{fig:architecture}
\end{figure}

\subsection{Design of Building Blocks}
The zoom indicator should be large for a region only when there exists at least one object whose spatial support mostly lies within the region, and whose size is sufficiently small compared to the region. The reasoning is that we should zoom in to a region only when it substantially increases the chance of detection. For example, if an object is mostly outside the region, dividing the region further is unlikely to increase the chance of detecting that object. Similarly, if an object is large compared to the current region, the task of detecting this object should be handled by this region or its parents. In the latter case, further division of the region not only wastes computational resources, but also introduces false positives in the region proposals. Figure \ref{fig:zoom} shows common situations and the desirable behavior of the zoom indicator. 

The role of adjacency prediction is to detect one or multiple objects that overlap with the anchor region sufficiently by providing tight bounding boxes. The adjacency prediction should be aware of the search geometry induced by the zoom indicator. More specifically, the adjacency prediction should perform well on the effective anchor regions induced by the search algorithm. For this purpose we propose a training procedure that is aware of the adaptive search scheme (discussed in Section \ref{sec:imp}). On the other hand, its design should explicitly account for typical geometric configurations of objects that fall inside the region, so that the training can be performed in a consistent fashion. For this reason, we propose to make predictions based on a set of sub-region priors as shown in Figure \ref{fig:sub-region}. Note that we also include the anchor region itself as an additional prior. We make sub-region priors large compared to the anchor under the intuition that if an object is small, it is best to wait until the features extracted are at the right scale to make bounding box predictions.

\subsection{Implementation}
\label{sec:imp}
We implement our algorithm using the Caffe \cite{jia2014caffe} framework, utilizing the open source infrastructure provided by the Fast R-CNN repository \cite{girshick2015fast}. In this section we introduce the implementation details of our approach. We use the Fast R-CNN detector since it is a fast and accurate recent approach. Our method should in principle work for a broad class of object detectors that use region proposals.

We train a deep neural network as illustrated in Figure \ref{fig:architecture}. Note that in addition to the sub-region priors as shown in Figure \ref{fig:sub-region}, we also add the region itself as a special prior region, making in total $11$ adjacency predictions per anchor. For the convolutional layers, we use the VGG16 model \cite{simonyan2014very} pre-trained on ImageNet data. The fully-connected layers are on top of a region pooling layer introduced in \cite{girshick2015fast} which allows efficient sharing of convolutional layer features. 

The training is performed as a three-step procedure. First, a set of regions is sampled from the image. These samples should contain hard positive and negative examples for both the zoom indicator and the adjacency prediction. Finally, the tuples of samples and labels are used in standard stochastic gradient descent training. We now discuss how the regions are sampled and labeled, and the loss function we choose.

\begin{figure}[t]
\begin{center}
      \includegraphics[height=0.7in, width=1.1in]{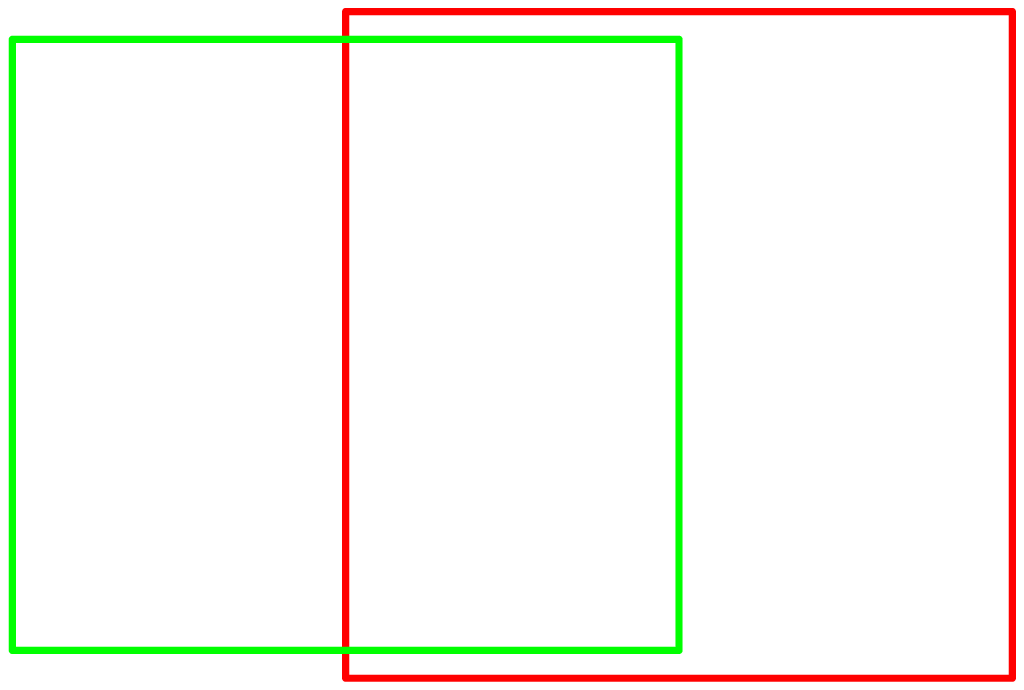}
      \hspace{1cm}
      \includegraphics[height=0.7in, width=0.8in]{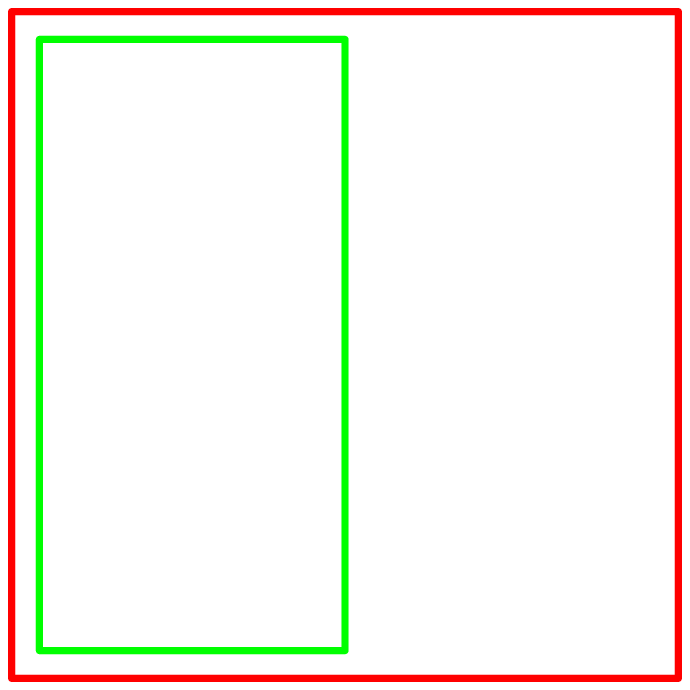} 
\end{center}
   \caption{Illustration of the inverse matching procedure. The red box is the inverse match for the object (green box). The left figure shows inverse matching of a neighboring square, the right figure shows inverse matching of a vertical stripe. } 
\label{fig:inverse-matching}
\end{figure}

\subsubsection{Region Sampling and Labeling}
Since a typical image only has a few object instances, to provide sufficient positive examples for adjacency predictions our method inversely finds regions that will see a ground truth object as a perfect fit to its prior sub-regions (see Figure \ref{fig:inverse-matching} for illustration). This provides $k \times 11$ training examples for each image, where $k$ is the number of objects. 

To mine for negative examples and hard positive examples, we search the input image as in Algorithm \ref{alg:region-propose}. Note that the algorithm uses zoom indicators from the AZ-Net. Instead of optimizing AZ-Net with an on-policy approach (that uses the intermediate AZ-Net model to sample regions), which might cause training to diverge, we replace the zoom prediction with the zoom indicator label. However, we note that using the zoom label directly could cause overfitting, since at test time the algorithm might encounter situations where a previous zoom prediction is wrong. To improve the robustness of the model, we add noise to the zoom label by flipping the ground truth with a probability of $0.3$. We found that models trained without random flipping are significantly less accurate. For each input image we initiate this procedure with five sub-images and repeat it multiple times. We also append horizontally flipped images to the dataset for data augmentation. 

Assignment of labels for the zoom indicator follows the discussion of Section \ref{sec:design}. The label is 1 if there exists an object with 50\% of its area inside the region and the area is at most 25\% of the size of the region. Note that here we use a loose definition of inclusion to add robustness for objects falling between boundaries of anchors. For adjacency prediction, we set a threshold in the intersection-over-union (IoU) score between an object and a region. A region is assigned to detect objects with which it has sufficient overlap. The assigned objects are then greedily matched to one of the sub-regions defined by the priors shown in Figure \ref{fig:sub-region}. The priority in the matching is determined by the IoU score between the objects and the sub-regions. We note that in this manner multiple predictions from a region are possible. 

\subsubsection{Loss Function}
As shown in Figure \ref{fig:architecture}, the AZ-Net has three output layers. The zoom indicator outputs from a sigmoid activation function. To train it we use the cross-entropy loss function popular for binary classification. For the adjacency predictions, the bounding boxes are parameterized as in Fast R-CNN \cite{ren2015faster}. Unlike in Fast R-CNN, to provide multiple predictions from any region, the confidence scores are not normalized to a probability vector. Correspondingly we use smooth L1-loss for bounding box output and element-wise cross-entropy loss for confidence score output. The three losses are summed together to form a multi-task loss function. 

\subsubsection{Fast R-CNN Detectors}
The detectors we use to evaluate proposal regions are Fast R-CNN detectors trained using AZ-Net proposals. As in \cite{ren2015faster}, we implement two versions: one with unshared convolutional features and the other that shares convolutional features with AZ-Net. The shared version is trained using alternating optimization.

\begin{figure*}[t]
\begin{center}
   \includegraphics[width=1.75in]{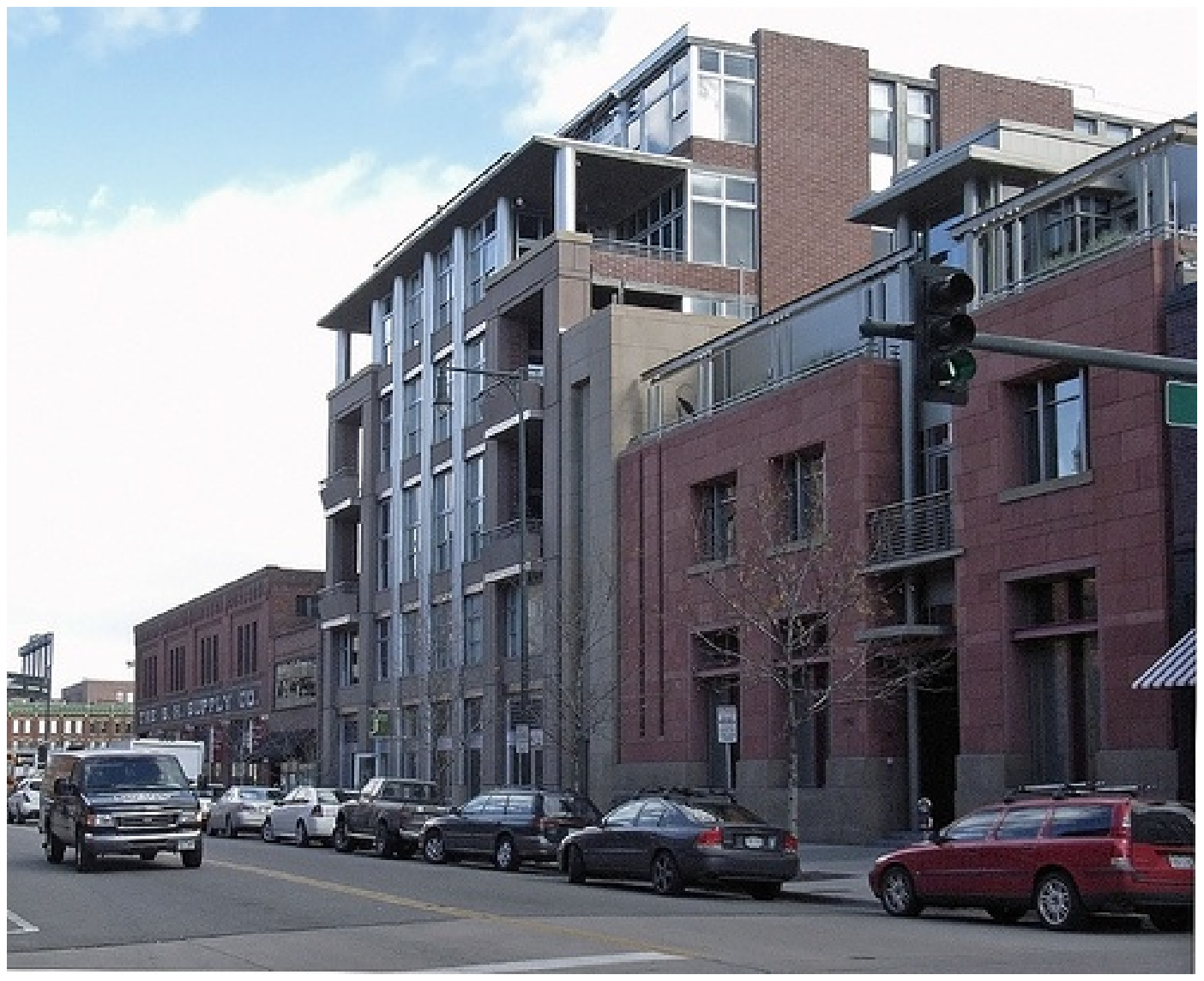}
   \includegraphics[width=1.75in]{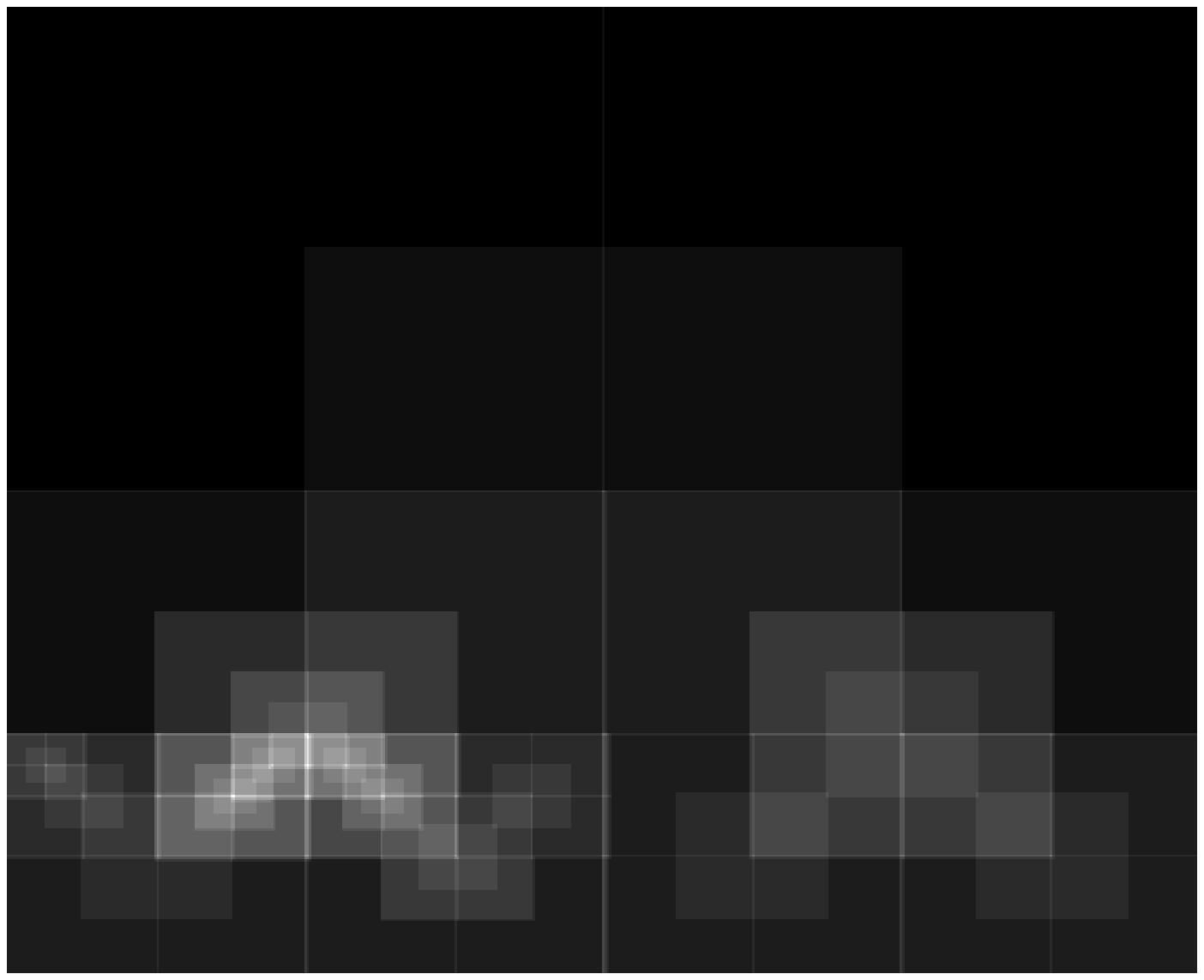}
   \includegraphics[width=1.75in]{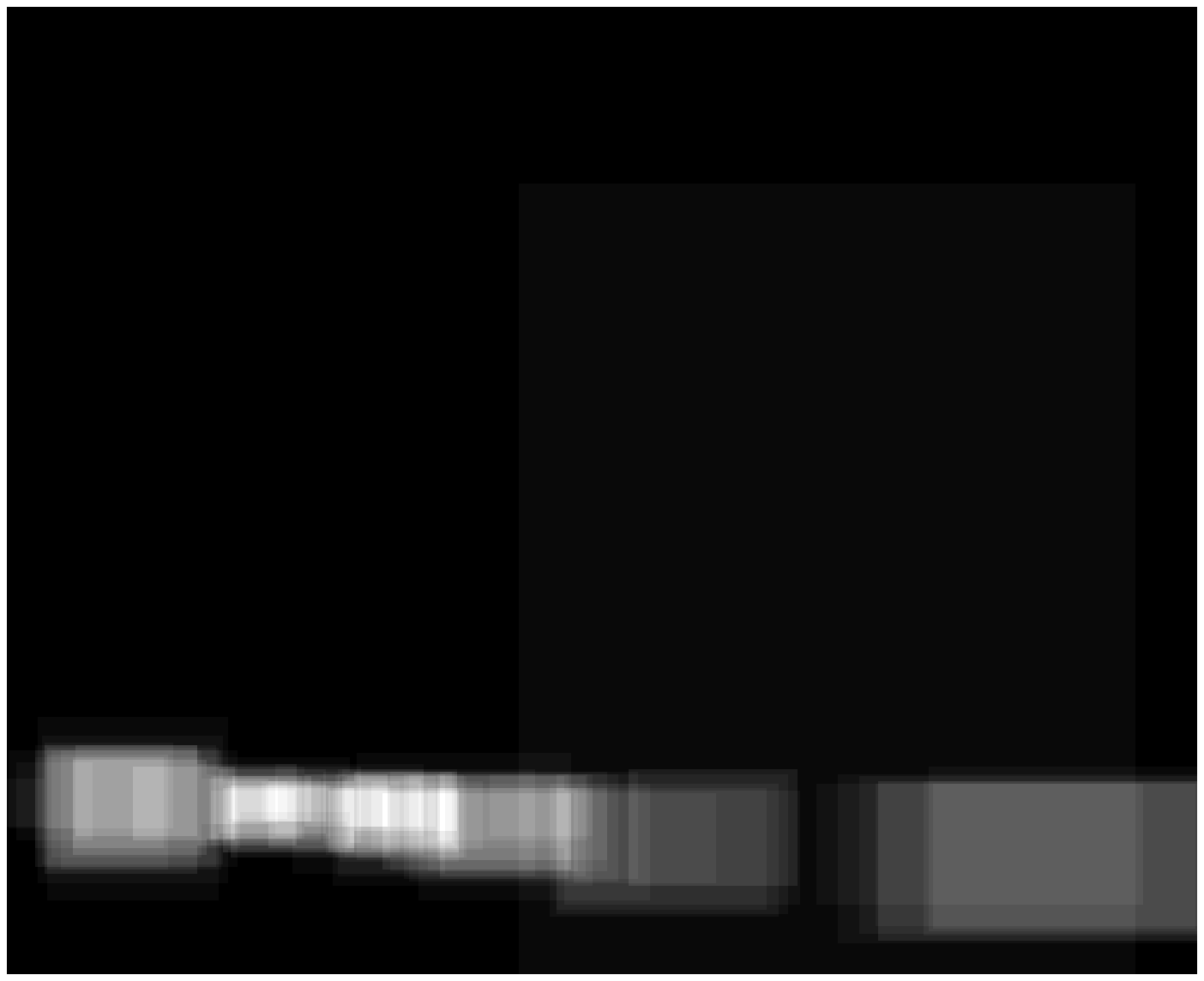}  \\   
   \includegraphics[width=1.75in]{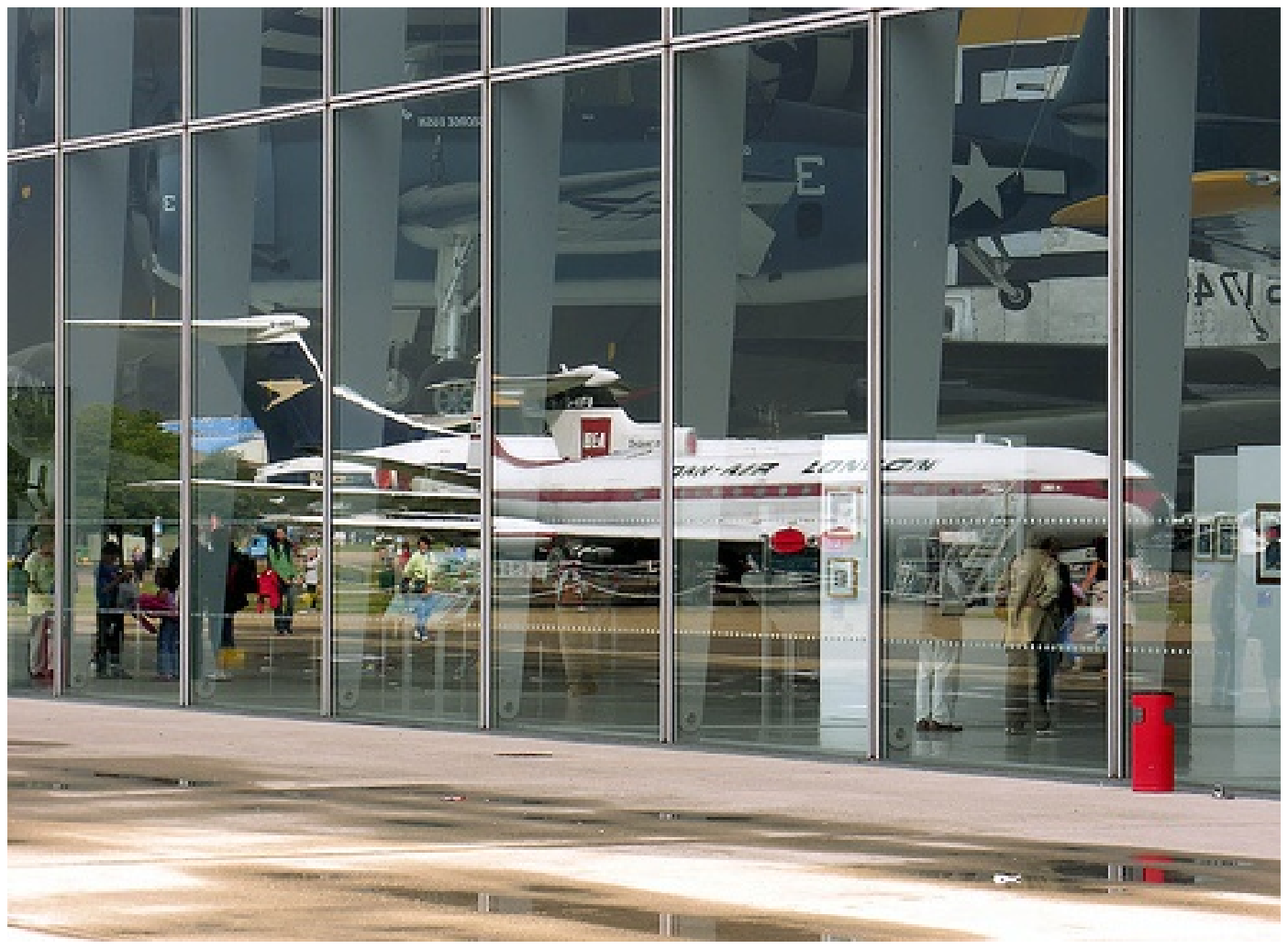}
   \includegraphics[width=1.75in]{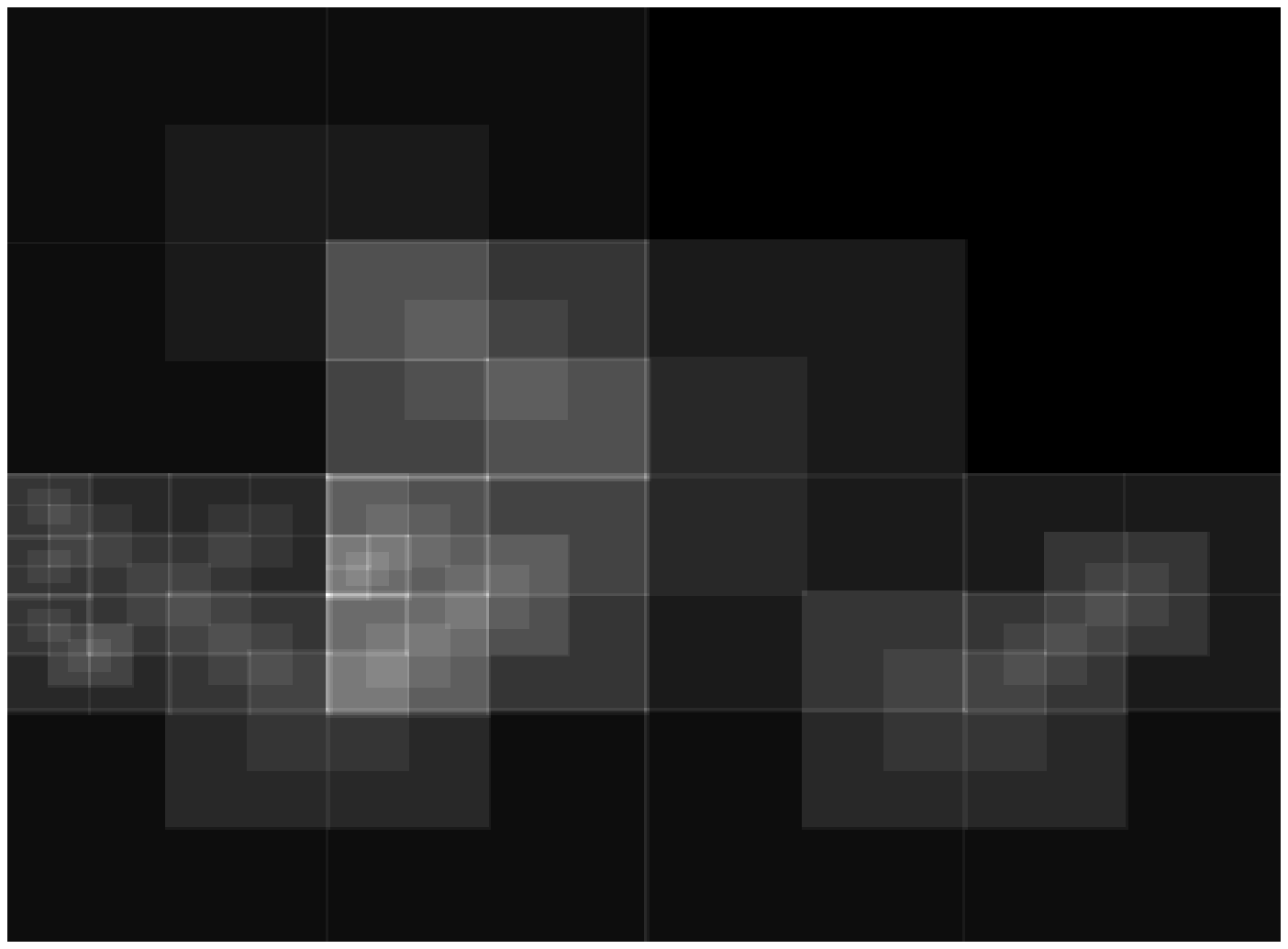}
   \includegraphics[width=1.75in]{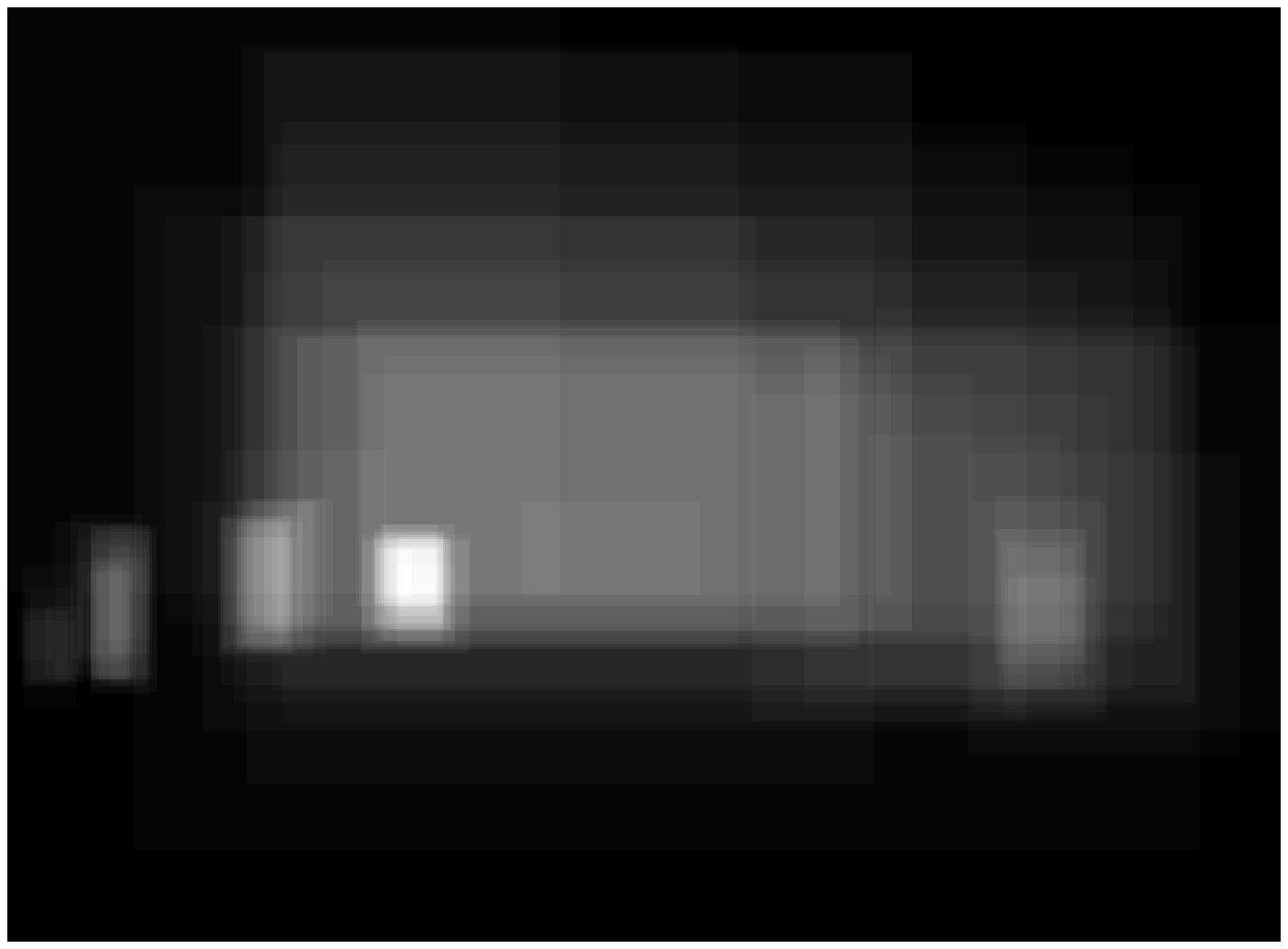} \\
   \includegraphics[width=1.75in]{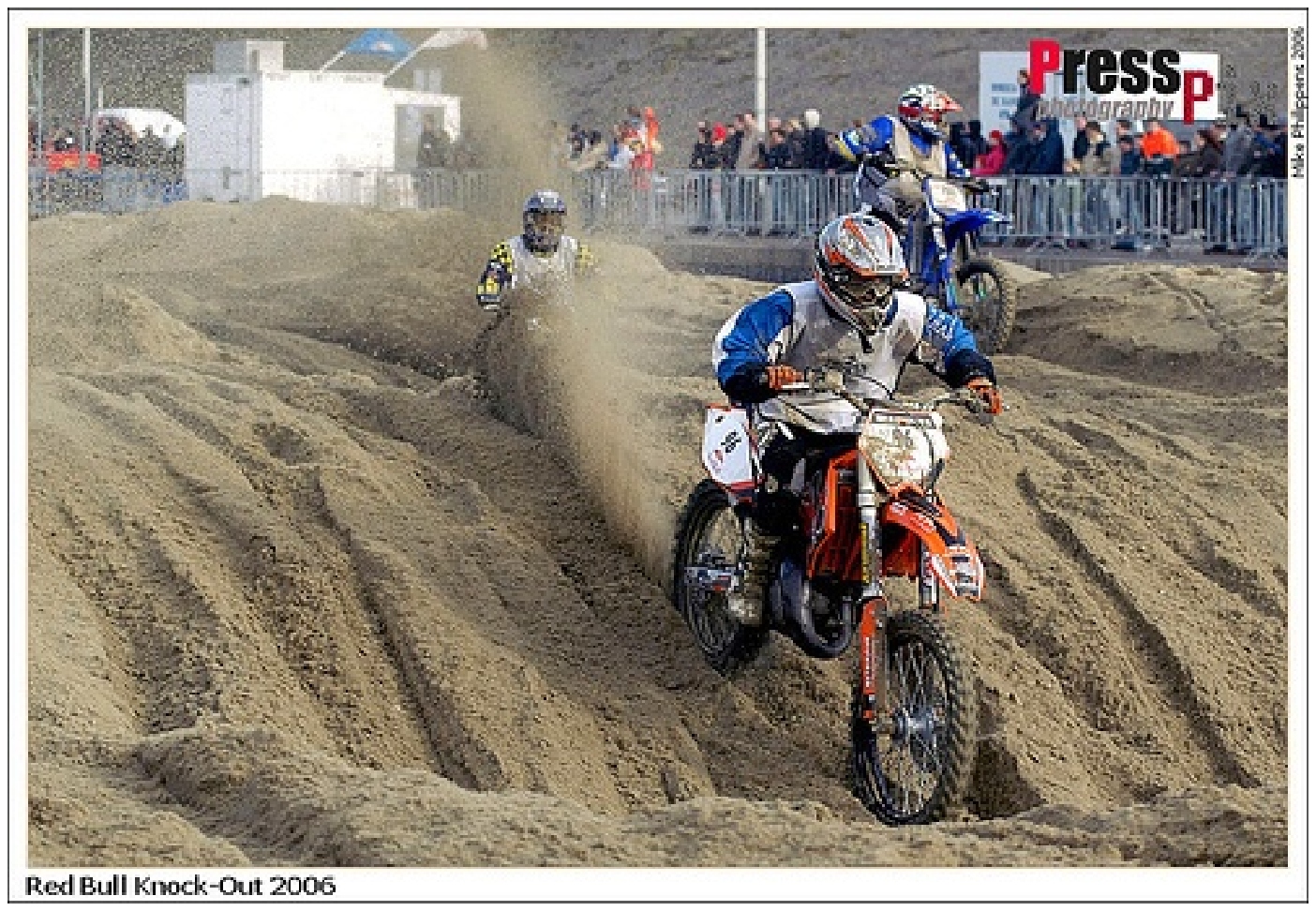}   
   \includegraphics[width=1.75in]{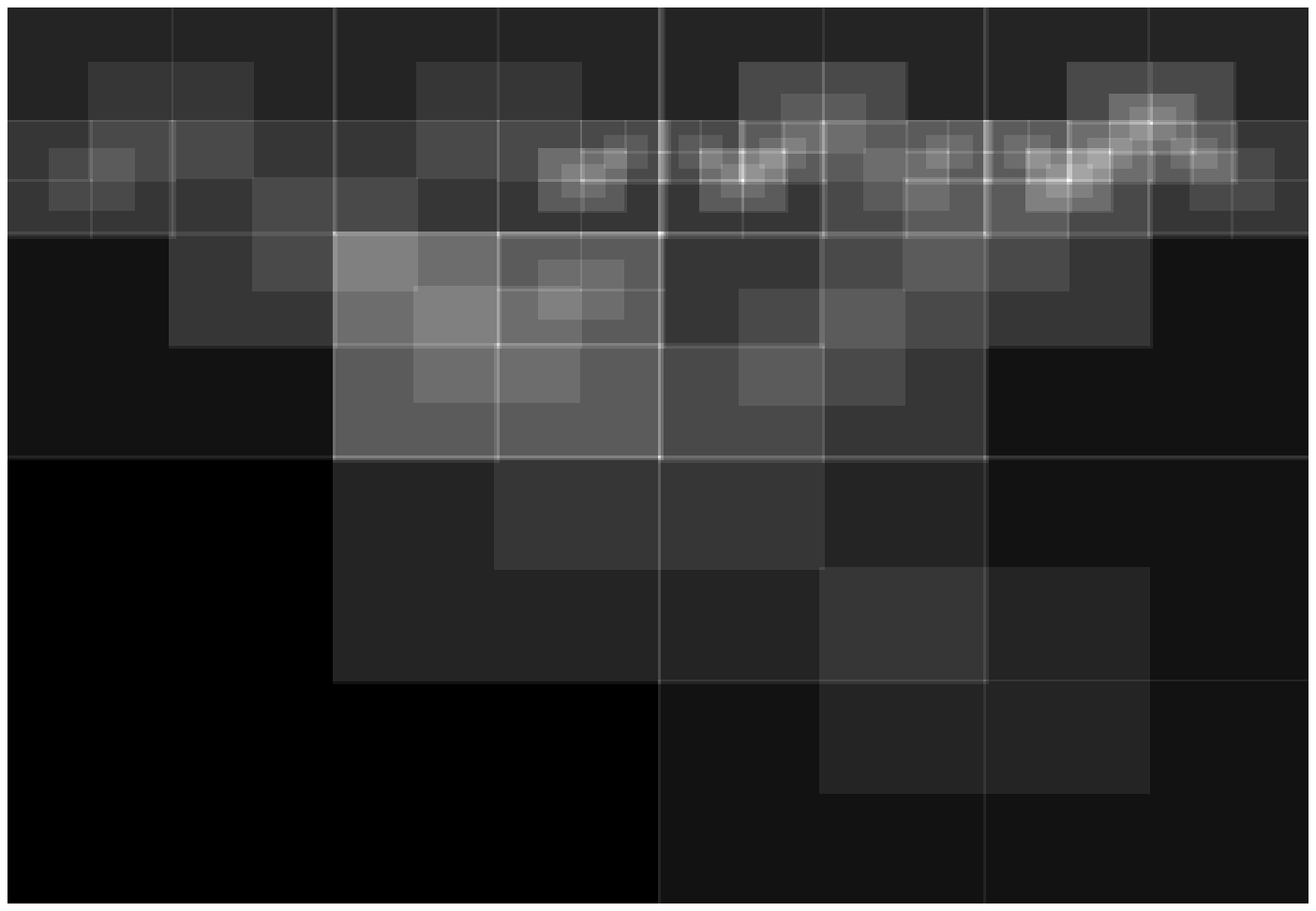}
   \includegraphics[width=1.75in]{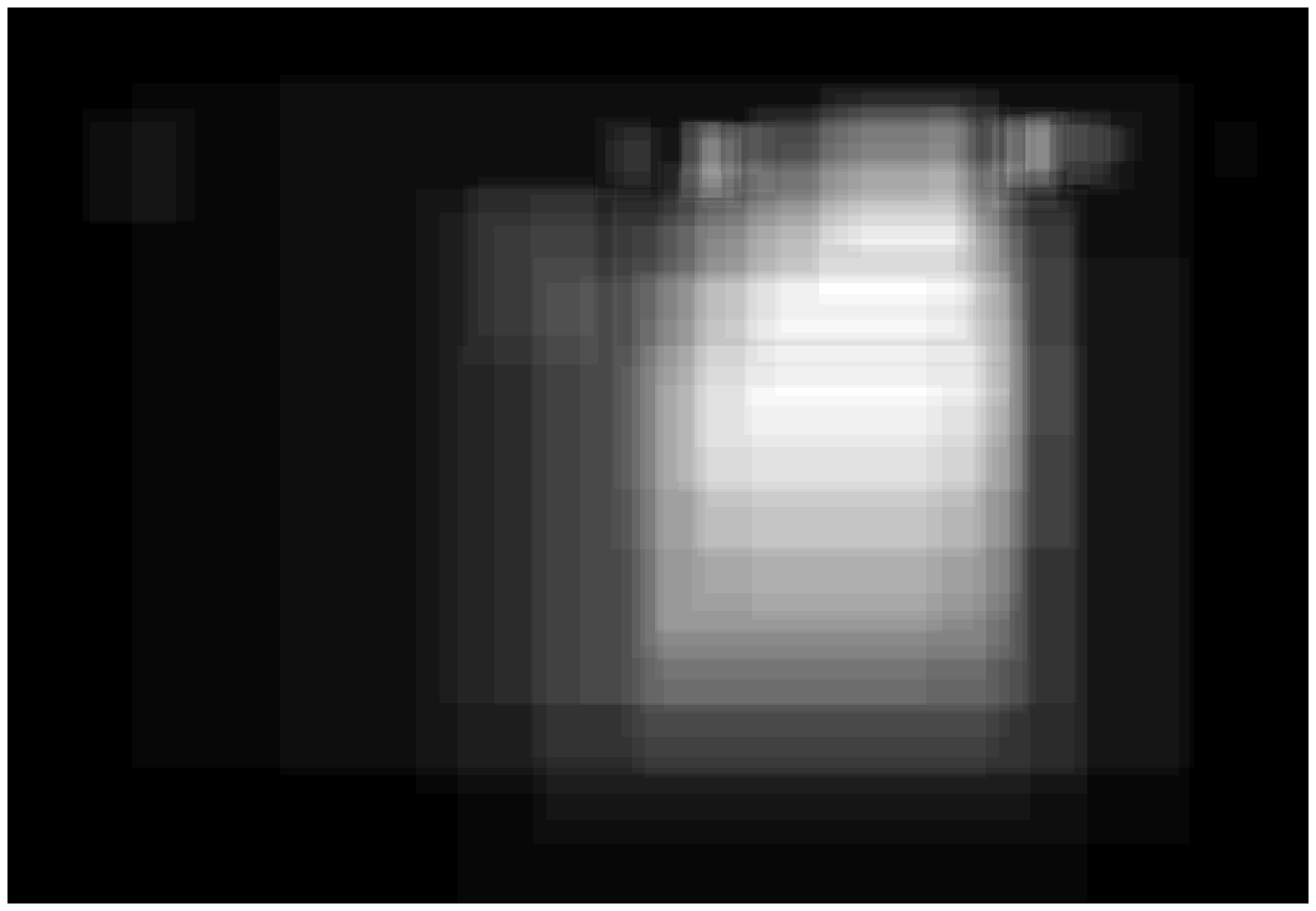}  \\
   \includegraphics[width=1.8in]{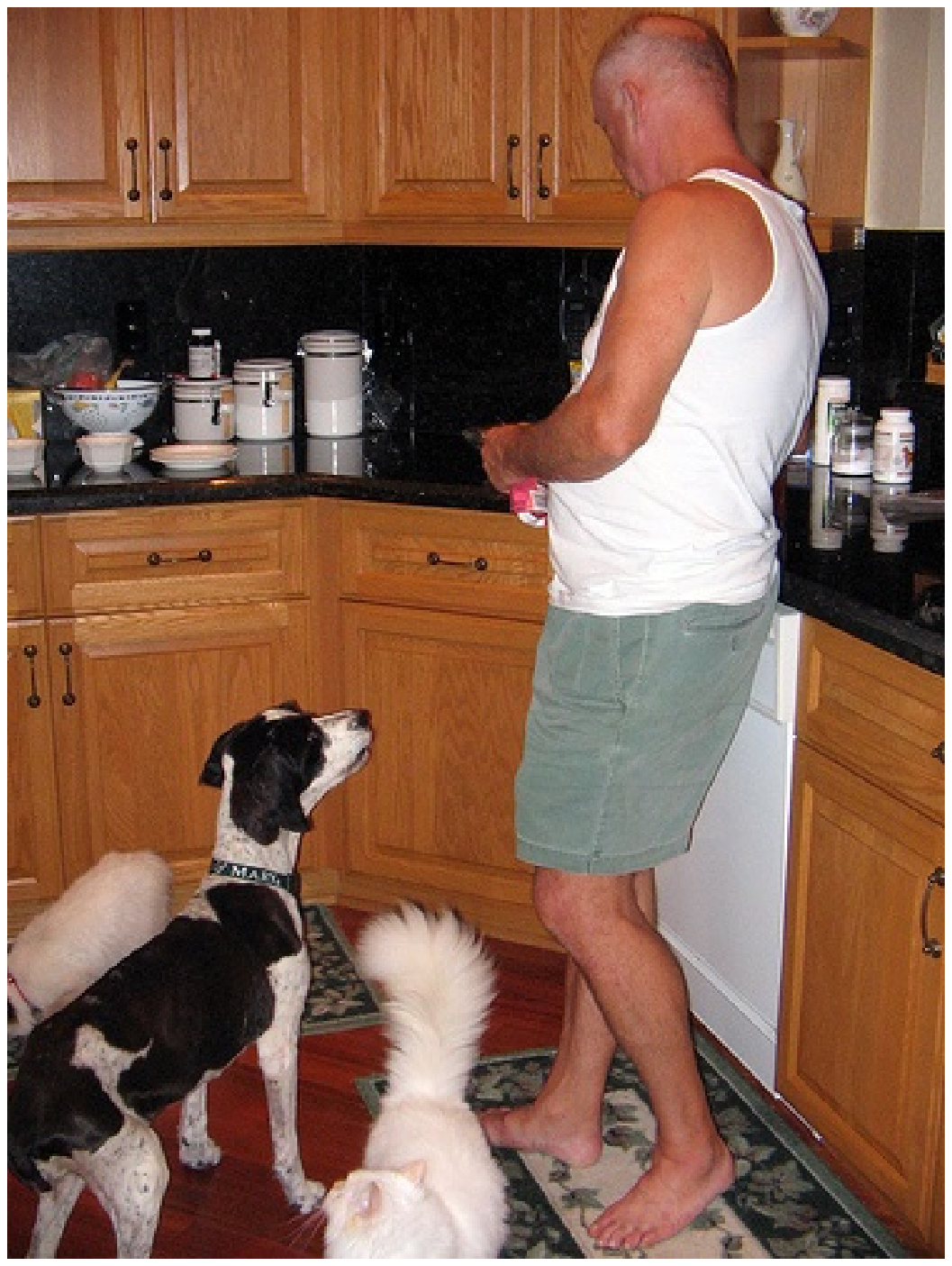}   
   \includegraphics[width=1.8in]{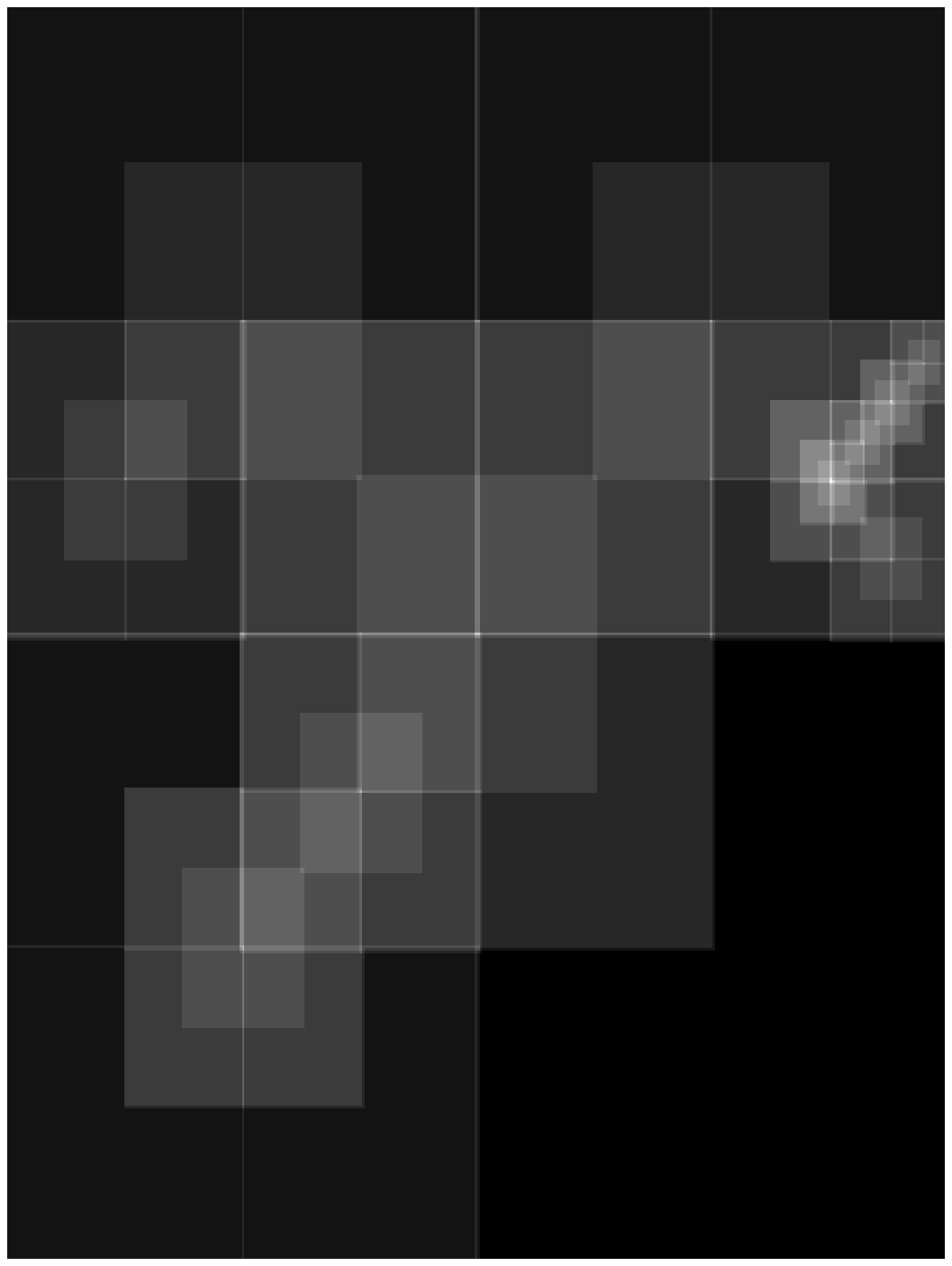}   
   \includegraphics[width=1.8in]{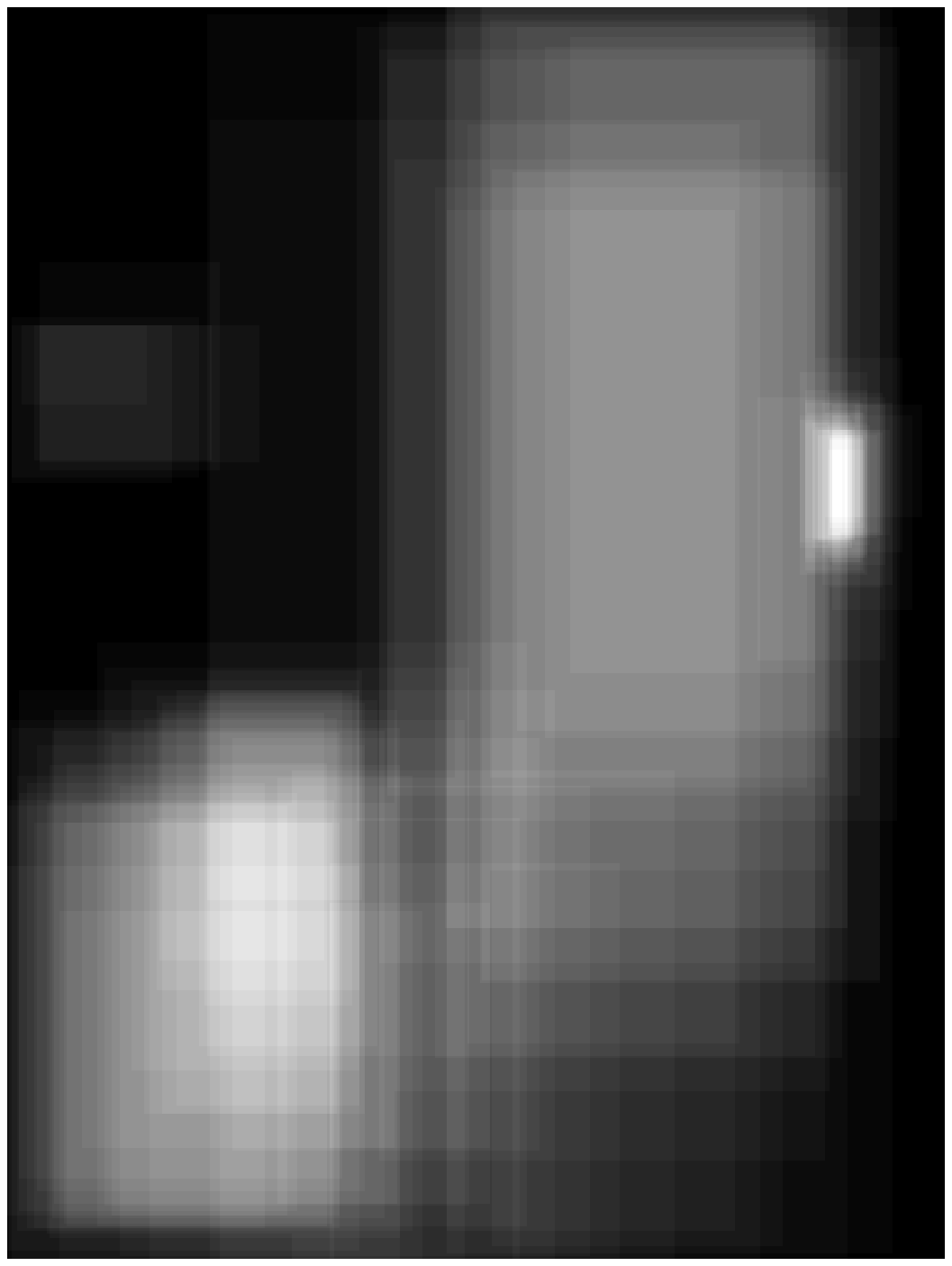}   
\end{center}
   \caption{Example outputs of our algorithm. The left column shows the original image. The middle column shows the anchor regions induced by our adaptive search. The right column shows the top 100 adjacency predictions made around the anchor regions. The anchor regions and the adjacency predictions are superimposed into a figure at the same resolution of the original image. We note that the anchor regions and the region proposals in our approach are shared across object categories. For example, for the last image, the algorithm generates anchor regions at proper scales near the dogs, the person, and the bottles.}
\label{fig:zoom_examples}
\end{figure*}

\section{Experiments}
\label{sec:exp}
We evaluate our approach on Pascal VOC 2007 \cite{pascal-voc-2007} and MSCOCO \cite{lin2014microsoft} datasets. In addition to evaluating the accuracy of the final detectors, we also perform detailed comparisons between the RPN approach adopted in Faster R-CNN and our AZ-Net on VOC 2007. At the end of the section, we give an analysis of the efficiency of our adaptive search strategy. 

\subsection{Results on VOC 2007}
To set up a baseline comparison, we evaluate our approach using the standard average precision (AP) metric for object detection. For AP evaluation we use the development kit provided by the VOC 2007 object detection challenge. We compare our approach against the recently introduced Fast R-CNN \cite{girshick2015fast} and Faster R-CNN \cite{ren2015faster} systems, which achieve state-of-the-art performance in standard benchmarks, such as VOC 2007 \cite{pascal-voc-2007} and VOC 2012 \cite{pascal-voc-2012}. A comparison is shown in Table \ref{tab:voc2007}. The results suggest that our approach is comparable to or better than these methods.

\begin{table*}
\tiny
\begin{center}
\begin{tabular}{|l|c|c|c|c|c|c|c|c|c|c|c|c|c|c|c|c|c|c|c|c|c|c|}
\hline
Method & boxes & mAP & aero & bike & bird & boat & bottle & bus & car & cat & chair & cow & table & dog & horse & mbike & person & plant & sheep & sofa & train & tv \\
\hline\hline
AZ-Net & 231 & 70.2 & 73.3 & 78.8 & 69.2 & \textbf{59.9} & 48.7 & \textbf{81.4} & 82.8 & 83.6 & 47.5 & 77.3 & 62.9 & \textbf{81.1} & 83.5 & 78.0 & 75.8 & 38.0 & 68.7 & 67.2 & 79.0 & 66.4 \\ 
AZ-Net* & 228 & \textbf{70.4} & 73.9 & 79.9 & 68.8 & 58.9 & 49.1 & 80.8 & \textbf{83.3} & \textbf{83.7} & 47.2 & 75.8 & 63.8 & 80.6 & \textbf{84.4} & \textbf{78.9} & 75.8 & \textbf{39.2} & 70.2 & 67.4 & 78.4 & \textbf{68.3}\\ \hline
RPN & 300 & 69.9 & 70.0 & \textbf{80.6} & \textbf{70.1} & 57.3 & 49.9 & 78.2 & 80.4 & 82.0 & \textbf{52.2} & 75.3 & \textbf{67.2} & 80.3 & 79.8 & 75.0 & \textbf{76.3} & 39.1 & 68.3 & 67.3 & \textbf{81.1} & 67.6\\
RPN* & 300 & 68.5 & 74.1 & 77.2 & 67.7 & 53.9  & \textbf{51.0} & 75.1 & 79.2 & 78.9 & 50.7 & \textbf{78.0} & 61.1 & 79.1 & 81.9 & 72.2 & 75.9 & 37.2 & \textbf{71.4} & 62.5 & 77.4 & 66.4\\ \hline
FRCNN & 2000 & 68.1 & \textbf{74.6} & 79.0 & 68.6 & 57.0 & 39.3 & 79.5 & 78.6 & 81.9 & 48.0 & 74.0 & 67.4 & 80.5 & 80.7 & 74.1 & 69.6 & 31.8 & 67.1 & \textbf{68.4} & 75.3 & 65.5 \\
\hline
\end{tabular}
\end{center}
\caption{Comparison on VOC 2007 test set using VGG-16 for convolutional layers. The results of RPN are reported in \cite{ren2015faster}. The results for Fast R-CNN are reported in \cite{girshick2015fast}. The AZ-Net and RPN results are reported for top-300 region proposals, but in AZ-Net many images have too few anchors to generate 300 proposals. * indicates results without shared convolutional features. 
All listed methods use DCNN models trained on VOC 2007 trainval. }
\label{tab:voc2007}
\end{table*}

\begin{figure}[t]
\begin{center}
   \includegraphics[width=3in]{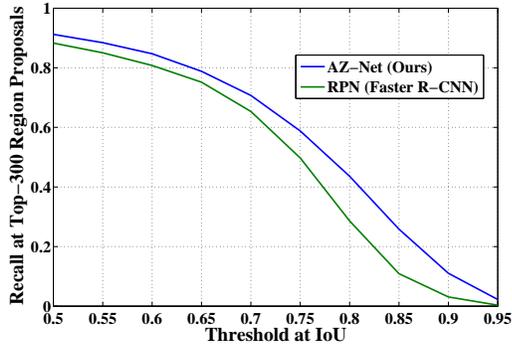}
\end{center}
   \caption{Comparison of recall of region proposals generated by AZ-Net and RPN at different intersection over union thresholds on VOC 2007 test. The comparison is performed at top-300 region proposals. Our approach has better recall at large IoU thresholds, which suggests that AZ-Net proposals are more accurate in localizing the objects. }
\label{fig:recall_thresh}
\end{figure}

\begin{figure}[t]
\begin{center}
   \includegraphics[width=2.5in]{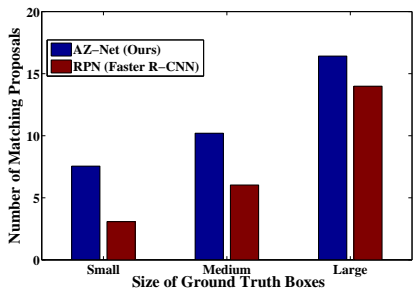}
\end{center}
   \caption{Number of proposals matched to ground truth (with IoU$=0.5$). This shows proposals from AZ-Net are more concentrated around true object locations.}
\label{fig:matching}
\end{figure}

\begin{figure}[t]
\begin{center}
   \includegraphics[width=3in]{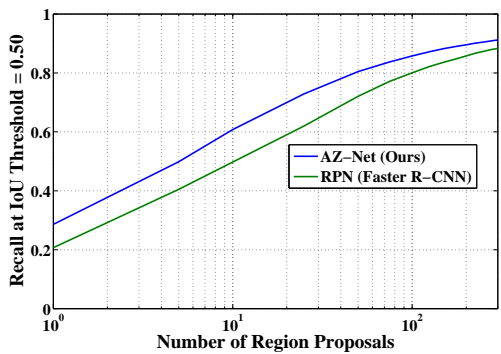}
\end{center}
   \caption{Comparison of recall of region proposals generated by AZ-Net and RPN at different number of region proposals on VOC 2007 test. The comparison is performed at IoU threshold $0.5$. Our approach has better early recall. In particular, it reaches $0.6$ recall with only 10 proposals. }
\label{fig:recall_early}
\end{figure}

\begin{figure}[t]
\begin{center}
   \includegraphics[width=2.5in]{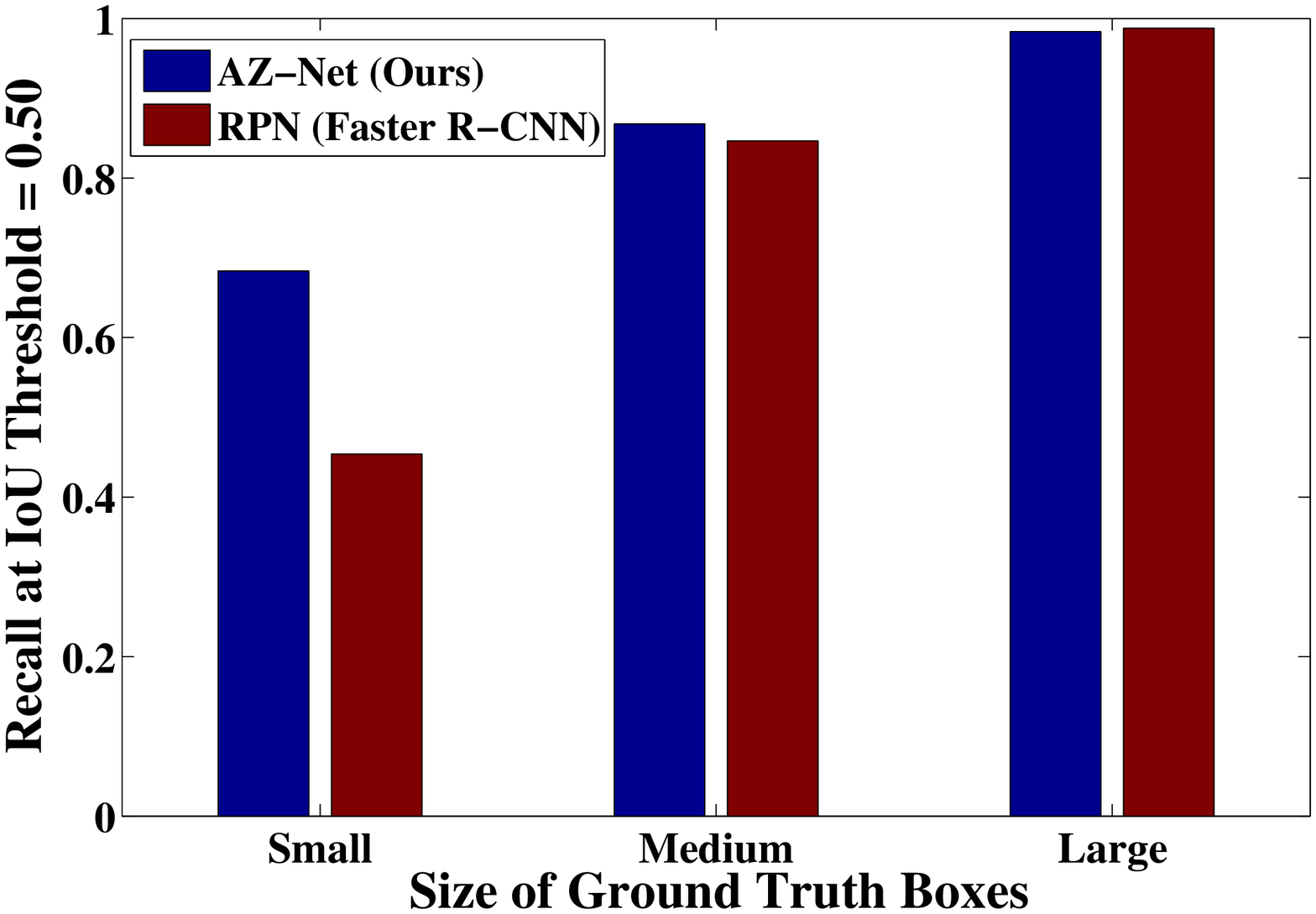}
\end{center}
   \caption{Comparison of recall of region proposals generated by AZ-Net and RPN for objects of different sizes on VOC 2007 test. The comparison is performed at IoU threshold $0.5$ with top-300 proposals. Our approach has significantly better recall for small objects.}
\label{fig:recall_area}
\end{figure}

\subsection{Quality of Region Proposals}
We preform a detailed analysis of the quality of region proposals from our AZ-Net, highlighting a comparison to the RPN network used in Faster R-CNN. For all our experiments, we analyze the recall on Pascal VOC 2007 test set using the following definition: An object is counted as retrieved if there exists a region proposal with an above-threshold IoU with it. The recall is then calculated as the proportion of the retrieved objects among all ground truth object instances. To accurately reproduce the RPN approach, we downloaded the region proposals provided on the Faster R-CNN repository \footnote{\url{https://github.com/ShaoqingRen/faster\_rcnn}}. We used the results from a model reportedly trained on VOC 2007 trainval. Correspondingly we compare it against our model trained on VOC 2007 trainval set. The comparisons concerning top-N regions are performed by ranking the region proposals in order of their confidence scores. 

Figure \ref{fig:recall_thresh} shows a comparison of recall at different IoU thresholds. Our AZ-net has consistently higher recall than RPN, and the advantage is larger at higher IoU thresholds. This suggests our method generates bounding boxes that in general overlap with the ground truth objects better. The proposals are also more concentrated around objects, as shown in Figure \ref{fig:matching}.

Figure \ref{fig:recall_early} shows a plot of recall as a function of the number of proposals. A region proposal algorithm is more efficient in covering objects if its area under the curve is larger. Our experiment suggests that our AZ-Net approach has a better early recall than RPN. That means our algorithm in general can recover more objects with the same number of region proposals. 

Figure \ref{fig:recall_area} shows a comparison of recall for objects with different sizes. The ``small object'' has an area less than $32^2$, a ``medium object'' has an area between $32^2$ and $96^2$, and a ``large object'' has an area greater than $96^2$, same as the definition in MSCOCO \cite{lin2014microsoft}. Our approach achieves higher recall on the small object subset. This is because when small objects are present in the scene our adaptive search strategy generates small anchor regions around them, as shown in Figure \ref{fig:zoom_examples}. 

\subsection{Efficiency of Adaptive Search}
\begin{figure}[t]
\begin{center}
   \includegraphics[width=3.5in, height=2in]{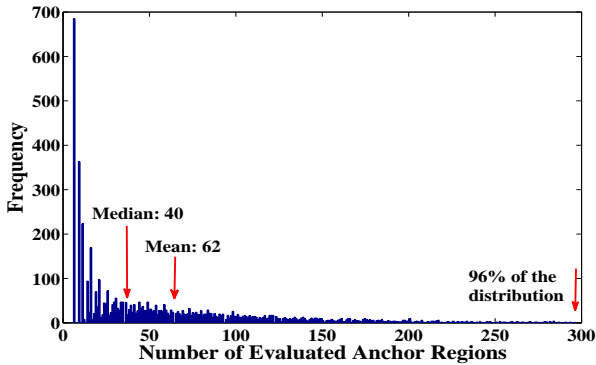}
\end{center}
   \caption{Distribution of the number of anchor regions evaluated on VOC 2007 test set. For most images a few dozen anchor regions are required. Note that anchors are shared across categories. }
\label{fig:anchor_freq}
\end{figure}

\begin{table}
\tiny
\begin{center}
\begin{tabular}{|l|c|c|c|}
\hline
Method & Anchor Regions & Region proposals & Runtime (ms) \\
\hline\hline
AZ-Net & 62 & 231 & 171 \\
AZ-Net* & 44 & 228 & 237 \\ \hline
RPN & 2400 & 300 & 198 \\
RPN* & 2400 & 300 & 342 \\ \hline
FRCNN & N/A & 2000 & 1830 \\
\hline
\end{tabular}
\end{center}
\caption{Numbers related to the efficiency of the object detection methods listed in Table \ref{tab:voc2007}. The runtimes for RPN and Fast R-CNN are reported for a K40 GPU \cite{ren2015faster}. Our runtime experiment is performed on a GTX 980Ti GPU. The K40 GPU has larger GPU memory, while the GTX 980Ti has higher clock rate. * indicates unshared convolutional feature version.}
\label{tab:runtime}
\end{table}

Our approach is efficient in runtime, as shown in Table \ref{tab:runtime}. We note that this is achieved even with several severe inefficiencies in our implementation. First, for each image our algorithm requires several rounds of fully connected layer evaluation, which induces expensive memory transfer between GPU and CPU. Secondly, the Faster R-CNN approach uses convolutional computation for the evaluation of anchor regions, which is highly optimized compared to the RoI pooling technique we adopted. Despite these inefficiencies, our approach still achieves high accuracy at a state-of-the-art frame rate, using lower-end hardware. With improved implementation and model design we expect our algorithm to be significantly faster. 

An interesting aspect that highlights the advantages of our approach is the small number of anchor regions to evaluate. To further understand this aspect of our algorithm, we show in Figure \ref{fig:anchor_freq} the distribution of anchor regions evaluated for each image. For most images our method only requires a few dozen anchor regions. This number is much smaller than the 2400 anchor regions used in RPN \cite{ren2015faster} and the 800 used in MultiBox \cite{erhan2014scalable}. Future work could further capitalize on this advantage by using an expensive but more accurate per-anchor step, or by exploring applications to very high-resolution images, for which traditional non-adaptive approaches will face intrinsic difficulties due to scalability issues. Our experiment also demonstrates the possibility of designing a class-generic search. Unlike per-class search methods widely used in previous adaptive object detection schemes \cite{caicedoactive,yoo2015attentionnet} our anchor regions are shared among object classes, making it efficient for multi-class detection. 

\begin{table}
\tiny
\begin{center}
\begin{tabular}{|l|c|c|}
\hline
Method & AP & AP IoU=0.50\\
\hline\hline
FRCNN (VGG16) \cite{girshick2015fast} & 19.7 & 35.9 \\ 
FRCNN (VGG16) \cite{ren2015faster} & 19.3 & 39.3 \\ \hline
RPN (VGG16) & 21.9 & 42.7 \\ 
RPN (ResNet) & 37.4 & 59.0 \\ \hline
AZ-Net (VGG16) & 22.3 & 41.0 \\
\hline
\end{tabular}
\end{center}
\caption{The detection mAP on MSCOCO 2015 test-dev set. The RPN (ResNet) entry won the MSCOCO 2015 detection challenge. Updated leaderboard can be found in \url{http://mscoco.org}.}
\label{tab:mscoco}
\end{table}

\subsection{Results on MSCOCO}
We also evaluated our method on MSCOCO dataset and submitted a ``UCSD'' entry to the MSCOCO 2015 detection challenge. Our post-competition work greatly improved accuracy with more training iterations. A comparison with other recent methods is shown in Table \ref{tab:mscoco}. Our model is trained with minibatches consisting of 256 regions sampled from one image, and 720k iterations in total. The results for RPN(VGG16) reported in \cite{ren2015faster} were obtained with an 8-GPU implementation that effectively has 8 and 16 images per minibatch for RPN and Fast R-CNN respectively, each trained at 320k training iterations. Despite the much shorter effective training iterations, our AZ-Net achieves similar mAP with RPN(VGG16) and is more accurate when evaluated on the MSCOCO mAP metric that rewards accurate localization.

Our best post-competition model is still significantly outperformed by the winning ``MSRA'' entry. Their approach is a Faster-R-CNN-style detection pipeline, replacing the VGG-16 network with an ultra-deep architecture called Deep Residual Network \cite{he2015deep}. They also report significant improvement from using model ensembles and global contextual information. We note that these developments are complementary to our contribution.

\section{Conclusion and Future Work}
\label{sec:con}
This paper has introduced an adaptive object detection system using adjacency and zoom predictions. Our algorithm adaptively focuses its computational resources on small regions likely to contain objects, and demonstrates state-of-the-art accuracy at a fast frame rate. 

The current method can be further extended and improved in many aspects. Better pre-trained models \cite{he2015deep} can be incorporated into the current system for even better accuracy. Further refining the model to allow single-pipeline detection that directly predicts class labels, as in YOLO \cite{redmon2015you} and the more recent SSD \cite{liu2015ssd} method, could significantly boost testing frame rate. Recent techniques that improve small object detection, such as the contextual model and skip layers adopted in Inside-Outside Net \cite{bell2015inside}, suggest additional promising directions. It is also interesting to consider more aggressive extensions. For instance, it might be advantageous to use our search structure to focus high-resolution convolutional layer computation on smaller regions, especially for very high-resolution images.

\subsection*{Acknowledgment} This work is supported by the National Science Foundation grants CIF-1302438, CCF-1302588 and CNS-1329819, as well as Xerox UAC, and the Sloan Foundation.

{\small
\bibliographystyle{ieee}
\bibliography{egbib}
}

\end{document}